\documentclass[smallcondensed]{svjour3}
\usepackage[utf8]{inputenc}

\usepackage[table,xcdraw]{xcolor}
\usepackage{graphicx}
\usepackage{subfig}
\usepackage{amsmath}
\usepackage{amsfonts}
\usepackage{chngpage}
\usepackage{multirow}
\usepackage{multicol}
\usepackage{makecell}
\usepackage{booktabs}
\usepackage{hyperref}

\hypersetup{
 colorlinks=true,
 linkcolor=blue,
 filecolor=blue,
 citecolor = black,      
 urlcolor=cyan,
}

\newcommand{\tcolA}{\textcolor{black}} 

\usepackage{natbib}

\newcommand{\nclusters}{k}
\newcommand{\nclusterslow}{\nclusters_{\text{L}}}
\newcommand{\nclustersmed}{\nclusters_{\text{M}}}
\newcommand{\nclustershigh}{\nclusters_{\text{H}}}

\begin{document}


\title{Interpretable Time Series Clustering Using Local Explanations}
\titlerunning{Interpretable Time Series Clustering Using Local Explanations}

\author{
        Ozan Ozyegen\and
        Nicholas Prayogo\and
       Mucahit Cevik \and
        Ayse Basar
}

\institute{Ozan Ozyegen \at
            Data Science Lab, Toronto Metropolitan University, Toronto, Canada\\
             \email{oozyegen@ryerson.ca}\\[-1.0em]
           \and
        Nicholas Prayogo \at 
          Data Science Lab, Toronto Metropolitan University, Toronto, Canada\\[-1.0em]
          \and
         Mucahit Cevik \at 
          Data Science Lab, Toronto Metropolitan University, Toronto, Canada\\[-1.0em]
          \and
          Ayse Basar\at 
          Data Science Lab, Toronto Metropolitan University, Toronto, Canada\\[-1.0em] 
}

\date{Received: date / Accepted: date}

\maketitle

\begin{abstract}
    This study focuses on exploring the use of local interpretability methods for explaining time series clustering models. 
    Many of the state-of-the-art clustering models are not directly explainable. 
    To provide explanations for these clustering algorithms, we train classification models to estimate the cluster labels. 
    Then, we use interpretability methods to explain the decisions of the classification models. 
    The explanations are used to obtain insights into the clustering models. 
    We perform a detailed numerical study to test the proposed approach on multiple datasets, clustering models, and classification models. 
    The analysis of the results shows that the proposed approach can be used to explain time series clustering models, specifically when the underlying classification model is accurate. 
    Lastly, we provide a detailed analysis of the results, discussing how our approach can be used in a real-life scenario. 
\end{abstract}

\section{Introduction}\label{intro_r3}

Clustering or cluster analysis is an unsupervised machine learning task that involves the automatic discovery of the natural grouping in data. 
The objects are grouped (i.e., clustered) based on how similar or different they are from those in other groups (i.e., clusters). 
Clustering can serve various purposes.
For instance, it can be highly useful for exploratory data analysis, where the underlying structure within the data can be identified by dividing the data into clusters. 

As the average cost of data storage and processing has decreased, data in many applications have started to be stored in the form of time series. 
Examples of such use-cases include historical prices of sales and stocks, exchange rates in finance, weather data, sensor measurements, and biometrics data. 
The availability of numerous time series datasets provides opportunities for many researchers to obtain insights through careful analysis of these datasets. 
Consequently, time series datasets have been used in various domains for a wide variety of purposes such as anomaly detection, indexing, clustering, classification, forecasting, visualization, segmentation, pattern detection, and trend analysis. 

Time series clustering is a special type of clustering method, where the clustered data is temporal. 
Time series data can be difficult to work with, e.g., due to naturally high data dimensionality and size. The clustering of these datasets can be particularly useful as it can lead to the discovery of interesting patterns. 
As a result, there are several research challenges to work with time series data.
For instance, it can be particularly important to develop methods that can recognize the dynamic changes in time series data for certain tasks such as anomaly detection, process control, and character recognition. 

As the machine learning models have found more and more use cases, the value of interpretability has started to be recognized. 
Interpretable AI, a recently developing machine learning field, consists of tools to make black box models more interpretable. 
Interpretability is important in many use cases of machine learning as it creates trust, transparency and fairness. 
When it is not understood why a model is making a decision, trusting the model can lead to inaccurate, and potentially dangerous decisions~\citep{caruana2017intelligible}. 
From the legal point of view, General Data Protection Regulation (GDPR) has introduced the ``right of explanation'' for all users in cases where an automated decision is made~\citep{marcinkevivcs2020interpretability}. 
Interpretability methods have started to be used heavily in the field of ``Responsible Artificial Intelligence''~\citep{marcinkevivcs2020interpretability}, to better satisfy such requirements. 

Not all AI models lack interpretability, some of the simpler ones are interpretable by design, although they are generally less accurate. 
Inspection of the model internals (e.g., parameters, architectural choices) are not always possible, and this is a problem with the majority of the machine learning algorithms. 
To remedy this issue, previous studies proposed a variety of tools to better understand how the model is working, and provide a reasonable explanation for the model's predictions. 
In time series clustering, interpretability is particularly challenging due to the high dimensionality and correlations of the features. 
Furthermore, it is difficult to find a standard approach, as different aspects of the dataset are important for different problems. 
For instance, in pricing problems, it is hard to find a match between the price increases and decreases of products. 
However, in other problems, it might be important to measure the stepwise similarity. 
The framework described in this paper can be used to interpret a variety of time series clustering models. 
It leverages the local interpretability methods and classification models that predict the cluster labels, to interpret the clustering models. 
This constitutes the main novelty of our study as, to the best of our knowledge, the time series classification models have not been previously used to interpret time series clustering methods.
The proposed approach is model-agnostic, and it enables an understanding of both the global importance of features in the dataset and the local importance of features for a specific clustering decision.

The main contributions of this paper can be summarized as follows:
\begin{itemize}
    \item 
    This is the first study that explores the use of model-agnostic local interpretability methods for explaining time series clustering models. 
    
    \item We perform detailed numerical experiments over multiple datasets, clustering models, classification models, and local interpretability methods. 
    We provide an elaborate discussion of the obtained results, comparing the accuracy and consistency of different methods. Additionally, we analyze both local and global explanations. 
    
    \item We demonstrate the usefulness of the proposed approach for an important practical problem in product pricing.
\end{itemize}

The rest of the paper is structured as follows. In Section~\ref{sec:related_work_r3}, we summarize the related work on interpretable time series clustering. 
In Section~\ref{sec:methodology_r3}, we describe the datasets, machine learning models, interpretability methods and experimental setup used in our analysis. 
In Section~\ref{sec:results_r3}, we analyze the performance of the clustering models, and the cluster prediction models, then examine the explanations generated by the local and global interpretability methods.
Finally, in Section~\ref{sec:conclusion_r3}, we summarize our findings and discuss potential future work.

\section{Related Work}\label{sec:related_work_r3}
In this section, we provide a summary of related works on interpretable time series clustering. 
Since our proposed methodology involves both clustering and classification methods, we make a distinction between the two, with a particular focus on time series modeling and interpretability. 

The majority of the time series clustering methods are not interpretable by design. 
K-means and k-medoids are two popular algorithms for time series clustering. 
These clustering algorithms can be trained using various distance metrics. 
Two well-known metrics suitable for time series are Euclidean distance, and dynamic time warping (DTW). 
DTW is typically considered to be superior to Euclidean distance, as it can measure the similarity between two temporal sequences which do not align exactly in speed, time, or length. 
However, the popular distance metrics do not work effectively for clustering all types of time series data.
For example, for the product pricing problem, where the products are clustered based on their price and sales patterns, \citet{bozanta2022time} proposed a ``movement pattern-based distance'', which allows effective clustering based on price information. 
Unlike most time series data, price information of products over time does not have a clear trend, seasonality, and cyclicity. 
However, similarity in direction of price movements, due to discounts, or corporate decisions is important to group the products based on their pricing behavior. 
Thus, depending on the problem and task at hand, researchers can also develop their custom distance metrics to handle certain types of datasets.
K-medioids can be preferred over k-means when custom distance metrics are used, as it can support any similarity measure. 
K-means, however, can only be used with distances that are constant with the mean, and it may fail to converge in certain cases. 
For example, absolute Pearson correlation is not consistent with the mean, and thus it can not be used with the k-means method. 

Since clustering methods are unsupervised learning algorithms, internal indices are typically used for assessing the performance of the clustering algorithms. 
A common internal measure for this purpose is the sum of squared distances of samples to their nearest cluster centers, also known as inertia. With this metric, a visual approach called the elbow method can be applied to show the number of clusters against inertia, and the number of clusters can then be selected based on where inertia starts to decrease at a visibly slower rate \citep{kodinariya2013review}. 
In another study, before training time series forecasting models on an electricity dataset, \citet{alvarez2010energy} apply k-means clustering to group the data, and they evaluate the internal validity using three well-known metrics, namely, Silhouette index, Davies-Bouldin index, and Dunn index. 
\citet{rousseeuw1987silhouettes} suggests using elbow method or silhouette method to find the optimal cluster size using silhouette index. 
In the elbow method, we compute the silhouette index for a range of candidate cluster sizes and pick the cluster size where the silhouette index falls sharply. 
In the silhouette method, we again compute the index for a range of candidate cluster sizes, but this time, we plot the silhouette index for each cluster separately. 
This allows the k-means clustering model to represent coefficients, and observe outliers and fluctuations within and between the clusters. 

As our methodology requires interpreting time series classification models that are trained to predict the cluster labels, we also review related work on interpretable time series classification. 
In general, we can divide these methods based on how the explanation is obtained, and the scope of the explanation~\citep{adadi2018peeking}. 
\textit{Intrinsically interpretable} models can sacrifice prediction performance to provide accurate and undistorted explanations. 
On the other hand, post-hoc methods are sometimes limited in their capacity to approximate the predictions of a complex model. 
In terms of scope, \textit{global interpretability} methods explain the entire reasoning of the model, whereas \textit{local interpretability} methods aim to explain the reasoning behind a specific decision. 

Time series classification has many real-life applications, and there is a wide variety of time series classification models in the literature~\citep{ismail2019deep}.
Machine learning models such as Random Forest and XGBoost are heavily used in Kaggle competitions for time series classification tasks, sometimes achieving the best performance~\citep{bojer2021kaggle}. 
Some of these machine learning models are interpretable. 
For example, XGBoost is based on the ensemble of decision tree classifiers, and the important features in the dataset can be found by computing the information gain brought by each feature. 
\citet{wang2016time} use an interpretable deep neural network, called Fully Convolutional Network (FCN), for time series classification, which achieves similar predictive accuracy compared to other state-of-the-art approaches. 
The use of the global average pooling layer within their convolutional model allows the model to extract the Class Activations Map (CAM) from the model activations. 
By using CAM, we can achieve local interpretability, and obtain the contribution of each input feature for the specific labels. 

Other researchers have used various black-box neural network-based models such as Recurrent Neural Networks, Temporal Convolutional Networks, and Transformers~\citep{ismail2020benchmarking}. 
The black-box models can be interpreted via post-hoc interpretability methods.
\citet{ismail2020benchmarking} study saliency-based post-hoc interpretability methods on various neural network architectures. 
They find that when temporal and feature domains are combined in a multivariate time series, saliency methods break down in general, and they propose a two-step rescaling approach to improve the quality of the explanations. 
Multiple studies have found that SHAP~\citep{lundberg2017unified} performs the best on time series datasets compared to other perturbation-based model interpretability methods~\citep{schlegel2019towards, ozyegen2022evaluation}. 
\citet{schlegel2019towards} evaluate widely used interpretability methods on time series data and introduce new verification strategies. 
They show that SHAP works robustly for all models, but other interpretability methods such as DeepLIFT, LRP, and saliency maps work better with specific architectures. 

The availability of various clustering models, classification models, and interpretability methods have led researchers to experiment with a variety of methodologies to achieve interpretability in clustering tasks. 
\citet{bertsimas2021interpretable} propose an interpretable clustering method using mixed integer programming models formulated over tree-based clustering models. 
Their approach requires first running a hierarchical clustering method and employing the resulting assignments as class labels. 
The data is then fit into a classification tree, and decision paths leading to each cluster's leaves are used to provide an explanation of the variable importance. 
\citet{cannone2020explainable} take this two-step model-specific framework and turn it into a model-agnostic framework for interpreting clustering models. 
In the first step, they train a clustering model (e.g., k-means, and agglomerative clustering) on a suitable dataset.
Then, the obtained cluster labels are classified with various machine learning models such as Random Forest, Decision Trees, and Neural Networks. 
Finally, they use the local interpretability methods such as LIME~\citep{ribeiro2016should} and SHAP~\citep{lundberg2017unified} to explain the model decisions. 
They test this approach on various synthetic and real-life datasets and provide a detailed comparison of the post-hoc interpretability methods. 

\section{Methodology}\label{sec:methodology_r3}
\begin{figure}[!ht]      
    \centering
    \includegraphics[width=0.73\textwidth]{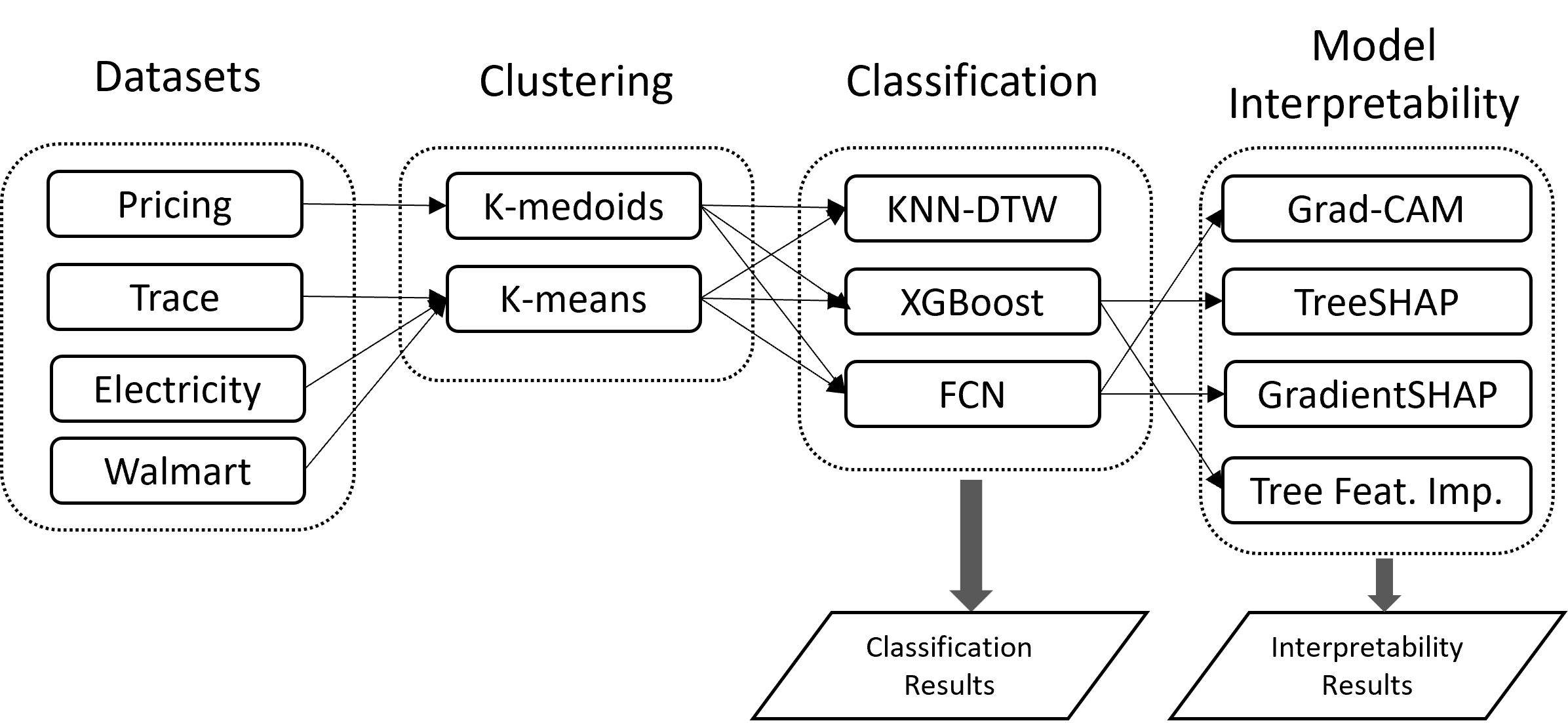}
    \caption{A flowchart of our proposed framework}
    \label{fig:methodology_r3}
\end{figure}

We next provide details on our proposed methodology by describing datasets, clustering and classification models, and local interpretability methods used in our analysis. 
Figure~\ref{fig:methodology_r3} provides an overview of our overall framework. 

\begin{figure}[!ht]      
    \centering
    \includegraphics[width=0.65\textwidth]{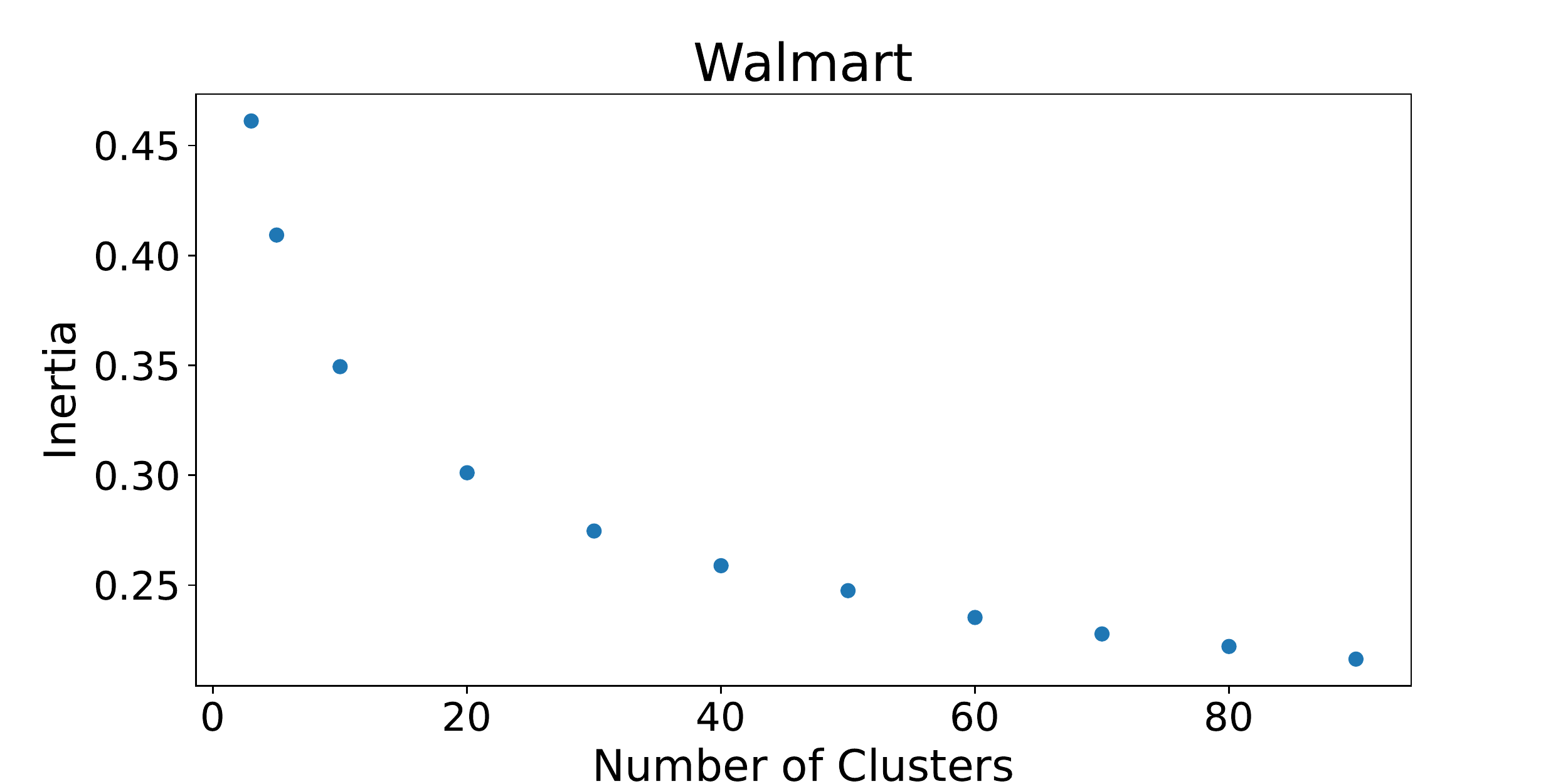}
    \caption{Elbow method with inertia for a representative dataset} 
    \label{fig:elbow_elect}
\end{figure}

The main idea is to use the classification models as a proxy to explain the decisions of clustering algorithms. 
We can divide the overall procedure into three steps. 
Firstly, we train a clustering algorithm and pick a number of clusters, $\nclusters$, suitable for the dataset at hand. 
Since there is no rule of thumb for $\nclusters$, we rely on internal validity metrics to identify a suitable value. 
Specifically, with inertia being the sum of squared distances of samples to their nearest cluster centers, we use the elbow method on the inertia curve of the clustering model, and we select the $\nclusters$ as the point where the rate of decrease of the inertia starts to visibly become much lower \citep{kodinariya2013review}. 
For instance, from Figure \ref{fig:elbow_elect}, the value for $\nclusters$ is chosen as 20, since the slope of the inertia curve starts to become visibly lower at this point. 
The values chosen from these plots are set as the medium number of clusters, $\nclustersmed$. 
Then, $\nclusterslow =  0.5\nclustersmed$ and $\nclustershigh =  2\nclustersmed$ are taken as the low and high number of clusters, respectively.

Below, we describe the datasets, clustering methods, classification models, and interpretability methods used in our analysis. 
We also describe the experimental setup and the hyperparameter tuning details at the end of this section. 

\subsection{Datasets}
We consider four distinct time series datasets in our analysis, namely, pricing, trace, Walmart and electricity, specifications of which are summarized in Table~\ref{tab:dataset_stats_r3}.

\setlength{\tabcolsep}{6pt}
\renewcommand{\arraystretch}{1.2}
\begin{table}[!htp]
\centering
\caption{Dataset specifications}\label{tab:dataset_stats_r3}
\begin{tabular}{|l|c|c|c|c|}
\hline
& \textbf{Pricing} & \textbf{Trace} & \textbf{Walmart} & \textbf{Electricity}\\ \hline
\# time series     & 1,160         & 200 & 2,660 & 370 \\ \hline
Domain             & $\mathbb{R}_+$ & $\mathbb{R}$ & $\mathbb{N}$ & $\mathbb{R}_+$ \\ \hline
\# timesteps & 571        &  275 & 143  & 168 \\ \hline
Time granularity     & Daily      & Secondly & Weekly & Hourly \\ \hline
\end{tabular} 
\end{table}

We provide more details on these datasets below.

\begin{itemize}\setlength\itemsep{0.3em}
\item \textit{Pricing dataset}:
The rapid growth and heavy competition in the grocery delivery industry have increased the importance of product pricing, and sales strategies for the companies.  
The clustering of similar price patterns can be used as a strategic tool and provide a competitive advantage to the retailers, by creating opportunities for better marketing and pricing strategies~\citep{bozanta2022time}. 
Interpretable clustering methodologies can improve the understanding of why certain products have similar pricing behavior, which can then be used strategically to create further competitive advantage. 
For instance, if certain products are priced together due to similar price changes during holidays, or special events, this information can be used to create a competitive pricing strategy. 
In this regard, we employ a dataset that contains the prices of 3,825 products over 570 days. 
It is obtained from Getir, an online food and grocery delivery company that operates in eight countries. 
Daily price values of the products are available as numerical values.
For the days where the price values are missing, we fill the missing values with the closest available price for the product. 
Also, we remove the products in which more than 80\% of the data points are not available. 
Next, we apply min-max scaling to each sample to put the range between 0.1 and 1. 
This is helpful as we aim to group products with similar price fluctuations, not based on their absolute price values.
After that, we analyze the pairwise distances between the time series and remove the outlier instances which are not likely to be clustered with any other instance. 
At the end of preprocessing steps, there remain 1,160 instances (i.e., products) in the pricing dataset. 

\item \textit{Trace dataset}:
This dataset is a part of the Transient Classification Benchmark~\citep{meesrikamolkul2012shape}. 
It is a synthetic dataset that simulates instrumentation failures in a nuclear power plant. 
The full dataset contains 150 instances, where 50 instances are available for each of the three classes. 
The instances are standardized, and they are linearly interpolated to have the same length of 275 data points. 
We do not fit a clustering model on the trace dataset, since the dataset was already separated into three classes, based on the similarity between the instances. 
The class labels of the trace dataset are used for classification to validate the quality of model explanations. 

\item \textit{Electricity dataset}:
The clustering of the electricity use of customers (i.e., households) is considered to be useful for energy companies in electricity distribution network planning, and load management~\citep{rasanen2009feature}. 
This dataset is part of UCI Machine Learning Repository~\citep{Dua:2019}, and it contains the hourly electricity consumption of 370 households over a four-year period. 
Since the input dimensionality is very high for a clustering task, and there is a high weekly seasonality in the dataset, we take a representative sample of one week (168 hours) from each household. 
We choose the time range between the start of 2013-05-01 to the end of 2013-05-07, and omit the time series with missing values, resulting in 336 time series for our analysis. 

\item \textit{Walmart dataset}:
The Walmart store sales forecasting dataset is released as part of a Kaggle competition in 2014\footnote{\url{https://www.kaggle.com/c/walmart-recruiting-store-sales-forecasting}}.
The dataset contains weekly store sales of 45 stores and 77 departments. 
Since this is a weekly dataset, we choose to use the entire time range from 2010-02-05 to 2012-10-26. We also treat each store-department combination as a separate time series, resulting in a total of 2,660 time series to be clustered.
\end{itemize}

\subsection{Clustering Methods}
We consider two specific clustering methods for grouping a given set of time series. 
\begin{itemize}\setlength\itemsep{0.3em}
    \item \textit{K-means}: K-means is a widely used clustering algorithm, which aims to minimize the within-cluster sum of squares (WCSS) measure, that is
    \begin{align*}
        W C S S=\sum_{i=1}^{k} \sum_{j=1}^{n}\left(x_{i j}-\mu_{i}\right)^{2}
    \end{align*}
    where $x_{ij}$ is the sample $j$ in the cluster $i$ and $\mu_i$ is the centroid (i.e., cluster center) of the cluster $i$. 
    Each sample is assigned to the nearest cluster center, and then the cluster centers are updated until convergence. 
    
    \item \textit{K-medoids}:
    Similar to k-means, the k-medoids algorithm aims to minimize the distance between points in a cluster, and a point designated as the center of that cluster. 
    However, unlike the k-means algorithm, k-medoids chooses actual data points as centers. 
    Furthermore, k-medoids can be used with arbitrary distance measures, whereas k-means is generally used with Euclidean distance, and it only supports certain distance measures. 
    We use the k-medoids method specifically for the pricing dataset since a custom distance metric is more suitable for this dataset. 
    Following prior work on clustering of pricing information~\citep{bozanta2022time}, we use the Movement Pattern-based Distance (MPBD) metric with this dataset. 
    MPBD is developed to capture similar pricing information of products. 
    When product prices change in opposite directions, the MPBD increases significantly, whereas an increase or decrease in a similar direction decreases the distances (i.e., increase the similarity) between the products. 
\end{itemize}

\subsection{Classification Models}\label{sec:classification_models_r3}
We test three feature configurations for classification model training: using the time series-only, using extracted-features-only, and using the time series with the extra features appended. 
In the time series-only setting, we feed the same data points to both clustering and classification models. 
For the other settings, we generate 20 extra features from the Time Series Feature Extraction Library (TSFEL) python package~\citep{barandas2020tsfel}. 
Across the three domains provided (i.e., statistical, temporal, and spectral), we choose to use their signal-based temporal domain features\footnote{Definition of each feature can be found at \url{https://tsfel.readthedocs.io/en/latest/descriptions/feature_list.html}} (e.g., autocorrelation and median difference) and add some common statistical features, namely, max, min, mean, variance and standard deviation. 
In the extracted-features-only setting, we only use the 20 features constructed from raw data to train the classification models. 
In the final setting, we combine the raw features with the 20 constructed features to train the classification models. 

\begin{itemize}\setlength\itemsep{0.3em}
    \item \textit{K-nearest Neighbors}: K-nearest neighbors (KNN) is a supervised learning method, which uses the distance between the data points to make a prediction about the label of a data point/instance. The class label is assigned based on the majority vote, where the most frequently represented class around a given data point is used. 
    We choose the number of neighbors used for classification as $K=5$. 
    The distance metric used with the KNN algorithm is typically specific to the dataset and the classification task.
    We use Dynamic Time Warping (DTW) as the distance metric, since it is particularly suitable for time series data, due to its ability to measure the similarity between time series instances that do not align perfectly. 

    \item \textit{XGBoost}: XGBoost (XGB) is a classification model, in the form of a boosted tree algorithm. 
    It is an ensemble method, optimized via gradient descent. 
    It is widely used in Kaggle competitions' top-performing solutions and achieves better results than a wide variety of machine learning models in many cases~\citep{bojer2021kaggle}. 
    Furthermore, the model is considered interpretable, as we can obtain the global importance of the features in the dataset by computing the average loss reduction gained when using a feature for splitting. 

    \item \textit{Fully Convolutional Network}:
    Convolutional Neural Networks (CNNs) are widely used in image classification problems. 
    They can achieve high predictive performance by using a series of convolutional filters, which can extract complex features from images. 
    As the CNN-based models gained more popularity, researchers have started applying various CNN-based architectures to solve the time series classification problems. 
    The Fully Convolutional Network (FCN) is one of these architectures that is well suited for time series classification problems~\citep{wang2016time}. 
    Specifically, the weight sharing between the temporally close features makes FCN a suitable model for time series classification. 
    FCN consists of subsequent 1-dimensional filters applied on time series to extract complex patterns at each convolutional layer, followed by a softmax layer at the end to extract the class probabilities. 

\end{itemize}

\subsection{Interpretability Methods}
Once the classification models are trained to predict the cluster id of a time series, we can employ a post-hoc interpretability method to understand the model predictions. 
In this way, the classification models serve as a surrogate model to explain the clustering model's decisions. 
In particular, we consider SHAP and Grad-CAM in our analysis. 
While SHAP is a model agnostic method, Grad-CAM is specifically designed for CNN models. 
\begin{itemize}\setlength\itemsep{0.3em}
    \item \textit{SHAP}: 
    SHAP~\citep{lundberg2017unified} is a post-hoc and model agnostic interpretability method that can explain the decisions of a model by estimating the contribution of inputs to the model predictions. 
    It is considered to be a part of a larger family of additive feature attribution methods (e.g., LIME~\citep{ribeiro2016should}, DeepLIFT~\citep{deepliftshrikumar2017learning}), which learn a local linear model to explain the original complex model. 
    SHAP is the only local explanation method that satisfies three desirable properties: local accuracy, missingness, and consistency~\citep{lundberg2017unified}. 
    Previous studies also propose model-specific variants, which have a lower time complexity and can approximate the SHAP values. 
    GradientSHAP~\citep{lundberg2017unified} and TreeSHAP~\citep{lundberg2020local} are the proposed variants for neural networks, and tree-based models, respectively. 
    Accordingly, in our analysis, we use GradientSHAP to explain the decisions of the FCN models, and TreeSHAP to explain the decisions of the XGB models. 
    We use the \texttt{shap} library in Python~\citep{lundberg2017unified} in our implementations.

    \item \textit{Grad-CAM}: 
    Gradient-weighted Class Activation Mapping (Grad-CAM) assigns feature importance values to each neuron for a particular prediction by using the gradient signal flowing into the last convolutional layer of a CNN \citep{selvaraju2017grad}. The traditional CAM relies on a global average pooling layer preceding the final output layer of the CNN, however, Grad-CAM does not require such adjustments to be made on the network architecture, allowing it to still give interpretations on most CNN architectures. 
    For time series data where neighboring time steps might differ much more significantly than, for instance, neighboring pixels in an image, having an average pooling layer could remove crucial information (e.g., spikes) otherwise present in the consecutive timesteps. 
    Thus, Grad-CAM might be preferable for time series data. For instance, \citet{fauvel2021xcm} apply Grad-CAM to a CNN-based model trained on multivariate time series datasets and find that Grad-CAM can identify the important time series input features that impact the model predictions.
    Accordingly, we use Grad-CAM as another method to interpret the decisions of our FCN models.

\end{itemize}

\subsection{Experimental Setup}\label{sec:exp_details_r3}



For classification, we apply a 70-30 train-test split on each dataset-classification model pair. 
To reduce the total number of experiments, we tune the model parameters on the pairs with medium cluster sizes. 
Then, the models are retrained using the best parameters in terms of the accuracy metric, and they are evaluated on the test set. 
The hyperparameter tuning is performed using a grid search approach, where we consider the parameter values given in Table~\ref{tab:hyper_range_r3}. 
The parameters that maximize the accuracy metric for the medium cluster sizes are then used to train the dataset-classification model pairs with the low and high cluster sizes. 
For the FCN model, we experiment with different number of convolutional layers, filter sizes, learning rates, and optimizers that are used to update the weights. 
For the KNN model, we test different values for the number of neighbors that influence the prediction, and the distance metric used to find the closest neighbors. 
For the XGB model, we experiment with the gamma regularization parameters, which regularize the number of splits, and max depth parameters, which determine the length of the trees. 
We apply filtering on the clusters by removing clusters that have less than four samples. 
We also observe that there could be a significant imbalance between the clusters, and there can be one or two clusters with a very high number of time series. 
Accordingly, we downsample the largest cluster by preserving only $v$ samples, where $v$ is the mean number of samples per cluster. 
After the classification models are trained, the explanations are generated from the train set, as this part of the dataset is larger, and it is likely to provide a better representation of what the model has learned~\citep{ozyegen2022evaluation}. 

\renewcommand{\arraystretch}{1.23}
\begin{table}[!ht]
    \centering
    \caption{The hyperparameter tuning search space.}
    \label{tab:hyper_range_r3}
    \resizebox{0.65\textwidth}{!}{
    \begin{tabular}{lll}
    \toprule
    Model &  & Search space \\
    \midrule
    \multirow[t]{3}{*}{FCN} & & \makecell[l]{ \textit{optimizers}: \{Adam, SGD\},\\*[0.2em] \textit{learning rate}: \{0.01, 0.001, 0.0001\},\\*[0.2em]
    \textit{\# convolutional layers}: \{1,2,3,4\},\\*[0.2em]
    \textit{\# filters on 1st layer}: \{4, 16, 64, 128\}\\*[0.2em]
    }\\
    \midrule
    KNN & & \makecell[l]{\textit{\# of neighbors}: \{5, 10, 15\},\\*[0.2em] \textit{metric}: \{Minkowski, DTW\}}\\*[0.2em]
    \midrule
    XGB & & \makecell[l]{\textit{gamma}: \{0.0, 1.0, 2.0\},\\*[0.2em] \textit{max depth}: \{3,6,9\}}\\
    \bottomrule
    \end{tabular}
    }
\end{table}

We perform all experiments using a workstation with i9-9990K 3.6GHz CPU, RTX2070 SUPER 8Gb GPU, and 128Gb of RAM, with Debian Linux OS. 
We use \texttt{scikit-learn} library for k-means, k-medoids, and KNN implementations, \texttt{xgboost} library for XGB implementation, and \texttt{pytorch} library for FCN implementation. 

\section{Results}\label{sec:results_r3}
In this section, we first review the performance of the clustering and classification models. 
Then, we generate model explanations using multiple interpretability methods and analyze these explanations to gain insights into the model behavior.

\subsection{Clustering and Classification Performance}\label{sec:perf}
In Table~\ref{tab:clustering_results}, we evaluate the performance of the clustering models for a varying number of clusters. 
For the electricity and Walmart datasets, we use the k-means clustering algorithm. 
However, because the pricing dataset does not show common time series characteristics such as trend, seasonality and cyclicity, we use the k-medoids clustering algorithm with the custom MPBD metrics designed for clustering the pricing data~\citep{bozanta2022time}. 
We measure the clustering performance using four popular internal validity measures: silhouette score, Calinski-Harabasz (CH) score, Davies-Bouldin (DB) score, and inertia~\citep{bozanta2022time, cannone2020explainable}. 

\tcolA{Overall, we observe that the validity metrics  decrease in most cases, as the cluster sizes increase. }
The (medium) number of clusters, $\nclustersmed$, is taken as 20 for the electricity, and Walmart datasets, and 50 for the pricing dataset, as determined by the elbow method on the inertia metric. 
We also report the clustering and classification results for $\nclusterslow = 0.5\nclustersmed$ and $\nclustershigh = 2\nclustersmed$ to evaluate the performance of the proposed methodology under conditions where the clustering of the instances can be less accurate.

\begin{table}[!ht]
    \centering
    \caption{Clustering performance for different number of clusters}
    \label{tab:clustering_results}
    \resizebox{0.85\textwidth}{!}{
\resizebox{\textwidth}{!}{
\begin{tabular}{lcrrrr}
\toprule
\textbf{Dataset} & \textbf{\# Clusters} & \textbf{Silhouette} & \textbf{CH} & \textbf{DB} & \textbf{Inertia}\\
\midrule
Electricity & 10 & 0.144 & 59.64 & 1.733 & 1,107\\
Electricity & 20 & 0.104 & 40.42 & 1.524 & 854 \\
Electricity & 30 & 0.088 & 32.01 & 1.481 & 726 \\
Electricity & 40 & 0.075 & 26.84 & 1.537 & 646 \\
\midrule
Walmart & 10 & 0.112 & 340.93 & 2.204 & 8,096 \\
Walmart & 20 & 0.111 & 208.95 & 2.086 & 6,978 \\
Walmart & 30 & 0.102 & 158.30 & 2.198 & 6,364 \\
Walmart & 40 & 0.097 & 128.50 & 2.262 & 5,998 \\
\midrule
Pricing & 25 & -0.230 & 3.65 & 3.104 & 6,560 \\
Pricing & 50 & -0.310 & 3.79 & 2.692 & 6,175 \\
Pricing & 75 & -0.320 & 3.43 & 2.542 & 5,999 \\
Pricing & 100 & -0.309 & 3.27 & 2.424 & 5,711 \\
\bottomrule
\end{tabular}
}
    }
\end{table}

The classification results are shown in Table~\ref{tab:classification_electricity_r3} (electricity), Table~\ref{tab:classification_pricing_r3} (pricing), Table~\ref{tab:classification_trace_r3} (trace), and Table~\ref{tab:classification_walmart_r3} (Walmart). 
We use macro precision, recall, and F1 score to evaluate the classification performance of the models. 
For each dataset-model pair, we report the mean and standard deviation of these metrics over five uniquely seeded train-test splits. 
While we report results for three different cluster sizes (i.e., $4k_l=2k_M=k_H$), we particularly focus on the results for the medium cluster sizes, since the cluster size chosen using the elbow method is expected to have the most suitable separation between the clusters. 
For each cluster size, we report the number of clusters before and after the post-processing steps described in Section~\ref{sec:methodology_r3}. 
We also report results for three different feature configurations, which are described in Section~\ref{sec:classification_models_r3}. 

Firstly, comparing the classification models, we find that the XGB performs the best in terms of average F1 scores for most datasets and cluster sizes, followed by FCN and KNN-DTW models. 
The clear exception is the electricity dataset, where XGB performs significantly worse than the other two models in terms of average performance values for medium and high cluster sizes. 
Secondly, we compare the different feature configurations. 
Overall, we find that the configuration with the extra features (i.e., time series and extracted features) consistently leads to higher classification accuracy for the FCN model. 
For the XGB model, we see a similar pattern in the majority of the cases, and we see mixed results for the KNN-DTW model. 
The only exception is for the trace dataset, where the feature-only configuration performs best for multiple models, with a marginal improvement of less than 1\% in terms of average performance values. 
This is not a significant difference since the trace dataset contains much fewer samples than the other datasets, and misclassification of a single sample can lead to the observed difference. 
The consistent improvements in the accuracy for the FCN and XGB models are interesting since, even though the clustering models are not fitted using these extracted features, the use of these features in the classification models leads to a more accurate prediction of the cluster labels. 
When the extracted features are selected carefully, as highly predictive and interpretable features, they can make the model predictions more interpretable. 
For this reason, in addition to the ``time series and extracted feature'' configuration, we also train the classification models with the extracted-features-only configuration. 
Unfortunately, this configuration leads to worse classification accuracy in most cases, except for the trace dataset where KNN-DTW and FCN models perform better when only the extracted features are used. 
The higher performance on the trace dataset can be attributed to the lower number of clusters in this dataset. 
When the number of clusters and the complexity of the problem increase, we are more likely to lose critical information when only the extracted features are used. 
However, when the nature of the dataset is known, one can carefully construct features that are interpretable, and highly predictive for classifying the cluster labels. 
Consequently, the classification models can be trained only with these features, and the explanation can be based directly on these constructed features instead of the raw features. 

\setlength{\tabcolsep}{4.5pt} 
\renewcommand{\arraystretch}{1.13} 
\begin{table}[!ht]
    \centering
    \caption{Classification results for the Electricity dataset (ordered based on mean F1 score for each cluster size, performance values are reported as mean $\pm$ standard deviation over five random train-test splits)}
    \label{tab:classification_electricity_r3}
    \resizebox{1\textwidth}{!}{
    \begin{tabular}{lllrrrrr}
\toprule
\textbf{Dataset} & \textbf{Config.} & \textbf{Model} & \textbf{ $\nclusters^\dagger$} & \textbf{F1 score} & \textbf{Accuracy} & \textbf{Precision} & \textbf{Recall} \\
\midrule
Electricity &     with\_feats &   XGB &  10 $\rightarrow$ 9 &     0.863 $\pm$ 0.013 &     0.870 $\pm$ 0.007 &      0.872 $\pm$ 0.026 &   0.870 $\pm$ 0.007 \\
Electricity &         default &   KNN &  10 $\rightarrow$ 9 &     0.856 $\pm$ 0.030 &     0.862 $\pm$ 0.029 &      0.875 $\pm$ 0.027 &   0.862 $\pm$ 0.029 \\
Electricity &         default &   XGB &  10 $\rightarrow$ 9 &     0.855 $\pm$ 0.033 &     0.862 $\pm$ 0.025 &      0.865 $\pm$ 0.041 &   0.862 $\pm$ 0.025 \\
Electricity &     with\_feats &   FCN &  10 $\rightarrow$ 9 &     0.850 $\pm$ 0.030 &     0.856 $\pm$ 0.025 &      0.867 $\pm$ 0.025 &   0.856 $\pm$ 0.025 \\
Electricity &     with\_feats &   KNN &  10 $\rightarrow$ 9 &     0.849 $\pm$ 0.043 &     0.856 $\pm$ 0.039 &      0.870 $\pm$ 0.037 &   0.856 $\pm$ 0.039 \\
Electricity &         default &   FCN &  10 $\rightarrow$ 9 &     0.799 $\pm$ 0.058 &     0.808 $\pm$ 0.048 &      0.825 $\pm$ 0.044 &   0.808 $\pm$ 0.048 \\
Electricity &      feat\_only &   XGB &  10 $\rightarrow$ 9 &     0.635 $\pm$ 0.039 &     0.646 $\pm$ 0.036 &      0.660 $\pm$ 0.033 &   0.646 $\pm$ 0.036 \\
Electricity &      feat\_only &   KNN &  10 $\rightarrow$ 9 &     0.618 $\pm$ 0.032 &     0.640 $\pm$ 0.037 &      0.631 $\pm$ 0.041 &   0.640 $\pm$ 0.037 \\
Electricity &      feat\_only &   FCN &  10 $\rightarrow$ 9 &     0.586 $\pm$ 0.045 &     0.600 $\pm$ 0.040 &      0.614 $\pm$ 0.025 &   0.600 $\pm$ 0.040 \\
\midrule
Electricity &         default &   KNN & 20 $\rightarrow$ 15 &     0.789 $\pm$ 0.031 &     0.804 $\pm$ 0.030 &      0.811 $\pm$ 0.029 &   0.804 $\pm$ 0.030 \\
Electricity &     with\_feats &   KNN & 20 $\rightarrow$ 15 &     0.782 $\pm$ 0.024 &     0.800 $\pm$ 0.022 &      0.803 $\pm$ 0.022 &   0.800 $\pm$ 0.022 \\
Electricity &     with\_feats &   FCN & 20 $\rightarrow$ 15 &     0.782 $\pm$ 0.015 &     0.798 $\pm$ 0.016 &      0.800 $\pm$ 0.015 &   0.798 $\pm$ 0.016 \\
Electricity &         default &   FCN & 20 $\rightarrow$ 15 &     0.732 $\pm$ 0.051 &     0.749 $\pm$ 0.051 &      0.737 $\pm$ 0.054 &   0.749 $\pm$ 0.051 \\
Electricity &     with\_feats &   XGB & 20 $\rightarrow$ 15 &     0.711 $\pm$ 0.036 &     0.723 $\pm$ 0.040 &      0.720 $\pm$ 0.036 &   0.723 $\pm$ 0.040 \\
Electricity &         default &   XGB & 20 $\rightarrow$ 15 &     0.700 $\pm$ 0.020 &     0.715 $\pm$ 0.021 &      0.714 $\pm$ 0.027 &   0.715 $\pm$ 0.021 \\
Electricity &      feat\_only &   XGB & 20 $\rightarrow$ 15 &     0.592 $\pm$ 0.012 &     0.610 $\pm$ 0.006 &      0.601 $\pm$ 0.020 &   0.610 $\pm$ 0.006 \\
Electricity &      feat\_only &   FCN & 20 $\rightarrow$ 15 &     0.507 $\pm$ 0.016 &     0.515 $\pm$ 0.024 &      0.524 $\pm$ 0.021 &   0.515 $\pm$ 0.024 \\
Electricity &      feat\_only &   KNN & 20 $\rightarrow$ 15 &     0.474 $\pm$ 0.032 &     0.501 $\pm$ 0.035 &      0.478 $\pm$ 0.029 &   0.501 $\pm$ 0.035 \\
\midrule
Electricity &         default &   KNN & 40 $\rightarrow$ 22 &     0.705 $\pm$ 0.043 &     0.740 $\pm$ 0.042 &      0.728 $\pm$ 0.026 &   0.740 $\pm$ 0.042 \\
Electricity &     with\_feats &   KNN & 40 $\rightarrow$ 22 &     0.700 $\pm$ 0.031 &     0.733 $\pm$ 0.028 &      0.728 $\pm$ 0.033 &   0.733 $\pm$ 0.028 \\
Electricity &     with\_feats &   FCN & 40 $\rightarrow$ 22 &     0.604 $\pm$ 0.033 &     0.631 $\pm$ 0.024 &      0.629 $\pm$ 0.041 &   0.631 $\pm$ 0.024 \\
Electricity &         default &   FCN & 40 $\rightarrow$ 22 &     0.594 $\pm$ 0.087 &     0.616 $\pm$ 0.088 &      0.612 $\pm$ 0.071 &   0.616 $\pm$ 0.088 \\
Electricity &     with\_feats &   XGB & 40 $\rightarrow$ 22 &     0.545 $\pm$ 0.065 &     0.562 $\pm$ 0.061 &      0.556 $\pm$ 0.072 &   0.562 $\pm$ 0.061 \\
Electricity &         default &   XGB & 40 $\rightarrow$ 22 &     0.504 $\pm$ 0.085 &     0.529 $\pm$ 0.084 &      0.525 $\pm$ 0.087 &   0.529 $\pm$ 0.084 \\
Electricity &      feat\_only &   XGB & 40 $\rightarrow$ 22 &     0.408 $\pm$ 0.027 &     0.427 $\pm$ 0.037 &      0.418 $\pm$ 0.036 &   0.427 $\pm$ 0.037 \\
Electricity &      feat\_only &   FCN & 40 $\rightarrow$ 22 &     0.352 $\pm$ 0.032 &     0.371 $\pm$ 0.030 &      0.377 $\pm$ 0.026 &   0.371 $\pm$ 0.030 \\
Electricity &      feat\_only &   KNN & 40 $\rightarrow$ 22 &     0.340 $\pm$ 0.021 &     0.369 $\pm$ 0.014 &      0.366 $\pm$ 0.042 &   0.369 $\pm$ 0.014 \\
\bottomrule
\multicolumn{8}{l}{\small $\dagger$: \# of clusters before and after filtering out the clusters with low number of instances} \\
\end{tabular}
}
\end{table}

\setlength{\tabcolsep}{4.5pt} 
\renewcommand{\arraystretch}{1.13} 
\begin{table}[!ht]
    \centering
    \caption{Classification results for the Pricing dataset (ordered based on mean F1 score for each cluster size, performance values are reported as mean $\pm$ standard deviation over five random train-test splits)}
    \label{tab:classification_pricing_r3}
    \resizebox{1\textwidth}{!}{
    \begin{tabular}{lllrrrrr}
\toprule
\textbf{Dataset} & \textbf{Config.} & \textbf{Model} & \textbf{ $\nclusters^\dagger$} & \textbf{F1 score} & \textbf{Accuracy} & \textbf{Precision} & \textbf{Recall} \\
\midrule
Pricing &         default &   XGB &  25 $\rightarrow$ 16 &     0.892 $\pm$ 0.008 &     0.903 $\pm$ 0.008 &      0.895 $\pm$ 0.009 &   0.903 $\pm$ 0.008 \\
Pricing &     with\_feats &   XGB &  25 $\rightarrow$ 16 &     0.887 $\pm$ 0.020 &     0.899 $\pm$ 0.015 &      0.890 $\pm$ 0.023 &   0.899 $\pm$ 0.015 \\
Pricing &     with\_feats &   FCN &  25 $\rightarrow$ 16 &     0.881 $\pm$ 0.010 &     0.889 $\pm$ 0.009 &      0.888 $\pm$ 0.011 &   0.889 $\pm$ 0.009 \\
Pricing &         default &   FCN &  25 $\rightarrow$ 16 &     0.873 $\pm$ 0.015 &     0.879 $\pm$ 0.018 &      0.878 $\pm$ 0.012 &   0.879 $\pm$ 0.018 \\
Pricing &      feat\_only &   XGB &  25 $\rightarrow$ 16 &     0.714 $\pm$ 0.030 &     0.737 $\pm$ 0.028 &      0.727 $\pm$ 0.039 &   0.737 $\pm$ 0.028 \\
Pricing &     with\_feats &   KNN &  25 $\rightarrow$ 16 &     0.705 $\pm$ 0.010 &     0.744 $\pm$ 0.009 &      0.722 $\pm$ 0.009 &   0.744 $\pm$ 0.009 \\
Pricing &         default &   KNN &  25 $\rightarrow$ 16 &     0.684 $\pm$ 0.026 &     0.728 $\pm$ 0.025 &      0.729 $\pm$ 0.025 &   0.728 $\pm$ 0.025 \\
Pricing &      feat\_only &   FCN &  25 $\rightarrow$ 16 &     0.670 $\pm$ 0.025 &     0.682 $\pm$ 0.025 &      0.674 $\pm$ 0.030 &   0.682 $\pm$ 0.025 \\
Pricing &      feat\_only &   KNN &  25 $\rightarrow$ 16 &     0.631 $\pm$ 0.029 &     0.672 $\pm$ 0.029 &      0.621 $\pm$ 0.030 &   0.672 $\pm$ 0.029 \\
\midrule
Pricing &     with\_feats &   XGB &  50 $\rightarrow$ 29 &     0.893 $\pm$ 0.007 &     0.903 $\pm$ 0.005 &      0.900 $\pm$ 0.006 &   0.903 $\pm$ 0.005 \\
Pricing &     with\_feats &   FCN &  50 $\rightarrow$ 29 &     0.876 $\pm$ 0.010 &     0.882 $\pm$ 0.011 &      0.882 $\pm$ 0.007 &   0.882 $\pm$ 0.011 \\
Pricing &         default &   XGB &  50 $\rightarrow$ 29 &     0.872 $\pm$ 0.029 &     0.882 $\pm$ 0.027 &      0.881 $\pm$ 0.026 &   0.882 $\pm$ 0.027 \\
Pricing &         default &   FCN &  50 $\rightarrow$ 29 &     0.863 $\pm$ 0.016 &     0.872 $\pm$ 0.014 &      0.870 $\pm$ 0.017 &   0.872 $\pm$ 0.014 \\
Pricing &         default &   KNN &  50 $\rightarrow$ 29 &     0.684 $\pm$ 0.030 &     0.718 $\pm$ 0.028 &      0.700 $\pm$ 0.031 &   0.718 $\pm$ 0.028 \\
Pricing &     with\_feats &   KNN &  50 $\rightarrow$ 29 &     0.681 $\pm$ 0.022 &     0.716 $\pm$ 0.020 &      0.698 $\pm$ 0.021 &   0.716 $\pm$ 0.020 \\
Pricing &      feat\_only &   XGB &  50 $\rightarrow$ 29 &     0.660 $\pm$ 0.032 &     0.694 $\pm$ 0.031 &      0.665 $\pm$ 0.030 &   0.694 $\pm$ 0.031 \\
Pricing &      feat\_only &   FCN &  50 $\rightarrow$ 29 &     0.633 $\pm$ 0.021 &     0.647 $\pm$ 0.020 &      0.644 $\pm$ 0.026 &   0.647 $\pm$ 0.020 \\
Pricing &      feat\_only &   KNN &  50 $\rightarrow$ 29 &     0.568 $\pm$ 0.017 &     0.603 $\pm$ 0.022 &      0.551 $\pm$ 0.013 &   0.603 $\pm$ 0.022 \\
\midrule
Pricing &         default &   XGB & 100 $\rightarrow$ 43 &     0.816 $\pm$ 0.028 &     0.830 $\pm$ 0.026 &      0.830 $\pm$ 0.026 &   0.830 $\pm$ 0.026 \\
Pricing &     with\_feats &   XGB & 100 $\rightarrow$ 43 &     0.795 $\pm$ 0.013 &     0.810 $\pm$ 0.009 &      0.811 $\pm$ 0.015 &   0.810 $\pm$ 0.009 \\
Pricing &     with\_feats &   FCN & 100 $\rightarrow$ 43 &     0.781 $\pm$ 0.019 &     0.791 $\pm$ 0.018 &      0.794 $\pm$ 0.024 &   0.791 $\pm$ 0.018 \\
Pricing &         default &   FCN & 100 $\rightarrow$ 43 &     0.777 $\pm$ 0.035 &     0.788 $\pm$ 0.033 &      0.787 $\pm$ 0.032 &   0.788 $\pm$ 0.033 \\
Pricing &     with\_feats &   KNN & 100 $\rightarrow$ 43 &     0.604 $\pm$ 0.018 &     0.639 $\pm$ 0.016 &      0.623 $\pm$ 0.027 &   0.639 $\pm$ 0.016 \\
Pricing &         default &   KNN & 100 $\rightarrow$ 43 &     0.599 $\pm$ 0.017 &     0.630 $\pm$ 0.012 &      0.634 $\pm$ 0.027 &   0.630 $\pm$ 0.012 \\
Pricing &      feat\_only &   XGB & 100 $\rightarrow$ 43 &     0.566 $\pm$ 0.016 &     0.601 $\pm$ 0.015 &      0.582 $\pm$ 0.020 &   0.601 $\pm$ 0.015 \\
Pricing &      feat\_only &   FCN & 100 $\rightarrow$ 43 &     0.559 $\pm$ 0.023 &     0.569 $\pm$ 0.031 &      0.578 $\pm$ 0.031 &   0.569 $\pm$ 0.031 \\
Pricing &      feat\_only &   KNN & 100 $\rightarrow$ 43 &     0.482 $\pm$ 0.020 &     0.516 $\pm$ 0.021 &      0.478 $\pm$ 0.017 &   0.516 $\pm$ 0.021 \\
\bottomrule
\multicolumn{8}{l}{\small $\dagger$: \# of clusters before and after filtering out the clusters with low number of instances} \\
\end{tabular}
}
\end{table}

\setlength{\tabcolsep}{4.5pt} 
\renewcommand{\arraystretch}{1.13} 
\begin{table}[!ht]
    \centering
    \caption{Classification results for the Trace dataset (ordered based on mean F1 score for each cluster size, performance values are reported as mean $\pm$ standard deviation over five random train-test splits)}
    \label{tab:classification_trace_r3}
    \resizebox{1\textwidth}{!}{
    \begin{tabular}{lllrrrrr}
\toprule
\textbf{Dataset} & \textbf{Config.} & \textbf{Model} & \textbf{ $\nclusters^\dagger$} & \textbf{F1 score} & \textbf{Accuracy} & \textbf{Precision} & \textbf{Recall} \\
\midrule
Trace &      feat\_only &   FCN & 3 $\rightarrow$ 3 &     1.000 $\pm$ 0.000 &     1.000 $\pm$ 0.000 &      1.000 $\pm$ 0.000 &   1.000 $\pm$ 0.000 \\
Trace &      feat\_only &   KNN & 3 $\rightarrow$ 3 &     1.000 $\pm$ 0.000 &     1.000 $\pm$ 0.000 &      1.000 $\pm$ 0.000 &   1.000 $\pm$ 0.000 \\
Trace &      feat\_only &   XGB & 3 $\rightarrow$ 3 &     1.000 $\pm$ 0.000 &     1.000 $\pm$ 0.000 &      1.000 $\pm$ 0.000 &   1.000 $\pm$ 0.000 \\
Trace &     with\_feats &   XGB & 3 $\rightarrow$ 3 &     1.000 $\pm$ 0.000 &     1.000 $\pm$ 0.000 &      1.000 $\pm$ 0.000 &   1.000 $\pm$ 0.000 \\
Trace &     with\_feats &   FCN & 3 $\rightarrow$ 3 &     0.996 $\pm$ 0.010 &     0.996 $\pm$ 0.010 &      0.996 $\pm$ 0.009 &   0.996 $\pm$ 0.010 \\
Trace &     with\_feats &   KNN & 3 $\rightarrow$ 3 &     0.798 $\pm$ 0.044 &     0.813 $\pm$ 0.037 &      0.870 $\pm$ 0.037 &   0.813 $\pm$ 0.037 \\
Trace &         default &   XGB & 3 $\rightarrow$ 3 &     0.312 $\pm$ 0.044 &     0.316 $\pm$ 0.043 &      0.317 $\pm$ 0.043 &   0.316 $\pm$ 0.043 \\
Trace &         default &   KNN & 3 $\rightarrow$ 3 &     0.291 $\pm$ 0.070 &     0.307 $\pm$ 0.064 &      0.299 $\pm$ 0.063 &   0.307 $\pm$ 0.064 \\
Trace &         default &   FCN & 3 $\rightarrow$ 3 &     0.247 $\pm$ 0.051 &     0.271 $\pm$ 0.040 &      0.266 $\pm$ 0.060 &   0.271 $\pm$ 0.040 \\
\bottomrule
\multicolumn{8}{l}{\small $\dagger$: \# of clusters before and after filtering out the clusters with low number of instances} \\
\end{tabular}
}
\end{table}

\setlength{\tabcolsep}{4.5pt} 
\renewcommand{\arraystretch}{1.13} 
\begin{table}[!ht]
    \centering
    \caption{Classification results for the Walmart dataset (ordered based on mean F1 score for each cluster size, performance values are reported as mean $\pm$ standard deviation over five random train-test splits)}
    \label{tab:classification_walmart_r3}
    \resizebox{1\textwidth}{!}{
    \begin{tabular}{lllrrrrr}
\toprule
\textbf{Dataset} & \textbf{Config.} & \textbf{Model} & \textbf{ $\nclusters^\dagger$} & \textbf{F1 score} & \textbf{Accuracy} & \textbf{Precision} & \textbf{Recall} \\
\midrule
Walmart &     with\_feats &   XGB & 10 $\rightarrow$ 10 &     0.895 $\pm$ 0.010 &     0.895 $\pm$ 0.010 &      0.897 $\pm$ 0.009 &   0.895 $\pm$ 0.010 \\
Walmart &         default &   XGB & 10 $\rightarrow$ 10 &     0.871 $\pm$ 0.007 &     0.870 $\pm$ 0.007 &      0.875 $\pm$ 0.006 &   0.870 $\pm$ 0.007 \\
Walmart &     with\_feats &   FCN & 10 $\rightarrow$ 10 &     0.866 $\pm$ 0.021 &     0.866 $\pm$ 0.021 &      0.873 $\pm$ 0.020 &   0.866 $\pm$ 0.021 \\
Walmart &     with\_feats &   KNN & 10 $\rightarrow$ 10 &     0.838 $\pm$ 0.017 &     0.839 $\pm$ 0.017 &      0.843 $\pm$ 0.017 &   0.839 $\pm$ 0.017 \\
Walmart &         default &   KNN & 10 $\rightarrow$ 10 &     0.838 $\pm$ 0.014 &     0.838 $\pm$ 0.013 &      0.843 $\pm$ 0.014 &   0.838 $\pm$ 0.013 \\
Walmart &         default &   FCN & 10 $\rightarrow$ 10 &     0.838 $\pm$ 0.026 &     0.839 $\pm$ 0.026 &      0.845 $\pm$ 0.021 &   0.839 $\pm$ 0.026 \\
Walmart &      feat\_only &   XGB & 10 $\rightarrow$ 10 &     0.797 $\pm$ 0.008 &     0.799 $\pm$ 0.008 &      0.802 $\pm$ 0.007 &   0.799 $\pm$ 0.008 \\
Walmart &      feat\_only &   FCN & 10 $\rightarrow$ 10 &     0.777 $\pm$ 0.012 &     0.781 $\pm$ 0.011 &      0.784 $\pm$ 0.011 &   0.781 $\pm$ 0.011 \\
Walmart &      feat\_only &   KNN & 10 $\rightarrow$ 10 &     0.725 $\pm$ 0.013 &     0.728 $\pm$ 0.011 &      0.731 $\pm$ 0.012 &   0.728 $\pm$ 0.011 \\
\midrule
Walmart &     with\_feats &   XGB & 20 $\rightarrow$ 20 &     0.856 $\pm$ 0.013 &     0.856 $\pm$ 0.013 &      0.860 $\pm$ 0.012 &   0.856 $\pm$ 0.013 \\
Walmart &     with\_feats &   FCN & 20 $\rightarrow$ 20 &     0.829 $\pm$ 0.016 &     0.830 $\pm$ 0.016 &      0.840 $\pm$ 0.012 &   0.830 $\pm$ 0.016 \\
Walmart &         default &   XGB & 20 $\rightarrow$ 20 &     0.828 $\pm$ 0.017 &     0.828 $\pm$ 0.017 &      0.834 $\pm$ 0.016 &   0.828 $\pm$ 0.017 \\
Walmart &         default &   FCN & 20 $\rightarrow$ 20 &     0.811 $\pm$ 0.020 &     0.812 $\pm$ 0.020 &      0.822 $\pm$ 0.015 &   0.812 $\pm$ 0.020 \\
Walmart &     with\_feats &   KNN & 20 $\rightarrow$ 20 &     0.809 $\pm$ 0.012 &     0.811 $\pm$ 0.012 &      0.816 $\pm$ 0.012 &   0.811 $\pm$ 0.012 \\
Walmart &         default &   KNN & 20 $\rightarrow$ 20 &     0.806 $\pm$ 0.008 &     0.808 $\pm$ 0.007 &      0.813 $\pm$ 0.008 &   0.808 $\pm$ 0.007 \\
Walmart &      feat\_only &   XGB & 20 $\rightarrow$ 20 &     0.699 $\pm$ 0.021 &     0.705 $\pm$ 0.020 &      0.701 $\pm$ 0.020 &   0.705 $\pm$ 0.020 \\
Walmart &      feat\_only &   FCN & 20 $\rightarrow$ 20 &     0.674 $\pm$ 0.015 &     0.686 $\pm$ 0.010 &      0.690 $\pm$ 0.011 &   0.686 $\pm$ 0.010 \\
Walmart &      feat\_only &   KNN & 20 $\rightarrow$ 20 &     0.616 $\pm$ 0.010 &     0.626 $\pm$ 0.012 &      0.633 $\pm$ 0.014 &   0.626 $\pm$ 0.012 \\
\midrule
Walmart &     with\_feats &   XGB & 40 $\rightarrow$ 40 &     0.784 $\pm$ 0.011 &     0.785 $\pm$ 0.011 &      0.796 $\pm$ 0.009 &   0.785 $\pm$ 0.011 \\
Walmart &     with\_feats &   KNN & 40 $\rightarrow$ 40 &     0.773 $\pm$ 0.012 &     0.777 $\pm$ 0.011 &      0.785 $\pm$ 0.011 &   0.777 $\pm$ 0.011 \\
Walmart &     with\_feats &   FCN & 40 $\rightarrow$ 40 &     0.768 $\pm$ 0.021 &     0.772 $\pm$ 0.020 &      0.785 $\pm$ 0.016 &   0.772 $\pm$ 0.020 \\
Walmart &         default &   KNN & 40 $\rightarrow$ 40 &     0.767 $\pm$ 0.013 &     0.772 $\pm$ 0.013 &      0.779 $\pm$ 0.012 &   0.772 $\pm$ 0.013 \\
Walmart &         default &   FCN & 40 $\rightarrow$ 40 &     0.758 $\pm$ 0.014 &     0.763 $\pm$ 0.013 &      0.773 $\pm$ 0.013 &   0.763 $\pm$ 0.013 \\
Walmart &         default &   XGB & 40 $\rightarrow$ 40 &     0.729 $\pm$ 0.013 &     0.730 $\pm$ 0.014 &      0.743 $\pm$ 0.014 &   0.730 $\pm$ 0.014 \\
Walmart &      feat\_only &   XGB & 40 $\rightarrow$ 40 &     0.585 $\pm$ 0.013 &     0.590 $\pm$ 0.012 &      0.596 $\pm$ 0.013 &   0.590 $\pm$ 0.012 \\
Walmart &      feat\_only &   FCN & 40 $\rightarrow$ 40 &     0.566 $\pm$ 0.012 &     0.582 $\pm$ 0.010 &      0.595 $\pm$ 0.012 &   0.582 $\pm$ 0.010 \\
Walmart &      feat\_only &   KNN & 40 $\rightarrow$ 40 &     0.491 $\pm$ 0.015 &     0.504 $\pm$ 0.012 &      0.516 $\pm$ 0.020 &   0.504 $\pm$ 0.012 \\
\bottomrule
\multicolumn{8}{l}{\small $\dagger$: \# of clusters before and after filtering out the clusters with low number of instances} \\
\end{tabular}
}
\end{table}

\subsection{Interpretability Results}
We next present the model explanations produced by various interpretability methods and discuss how explanations can be used to gain insights into model behavior. 
We analyze the results in two ways. 
First, we look at the global explanations to analyze the overall importance of the features for the models. 
Then, we show the cluster explanations, which display the important features for a given cluster, along with the importance of the features for individual samples. 
It is worth noting that, using the SHAP method, we can find the contributions of the features to the prediction, instead of their importance to the predictions. 
However, to make it comparable with the other methods, we take the absolute values of the SHAP contribution scores and display the importance of the features. 

\subsubsection{Global Feature Importance}\label{sec:glob_feat_imp}
\begin{figure}[hbt!]
    \centering
        \subfloat[Pricing dataset 
        ]{\includegraphics[width=0.5\textwidth]{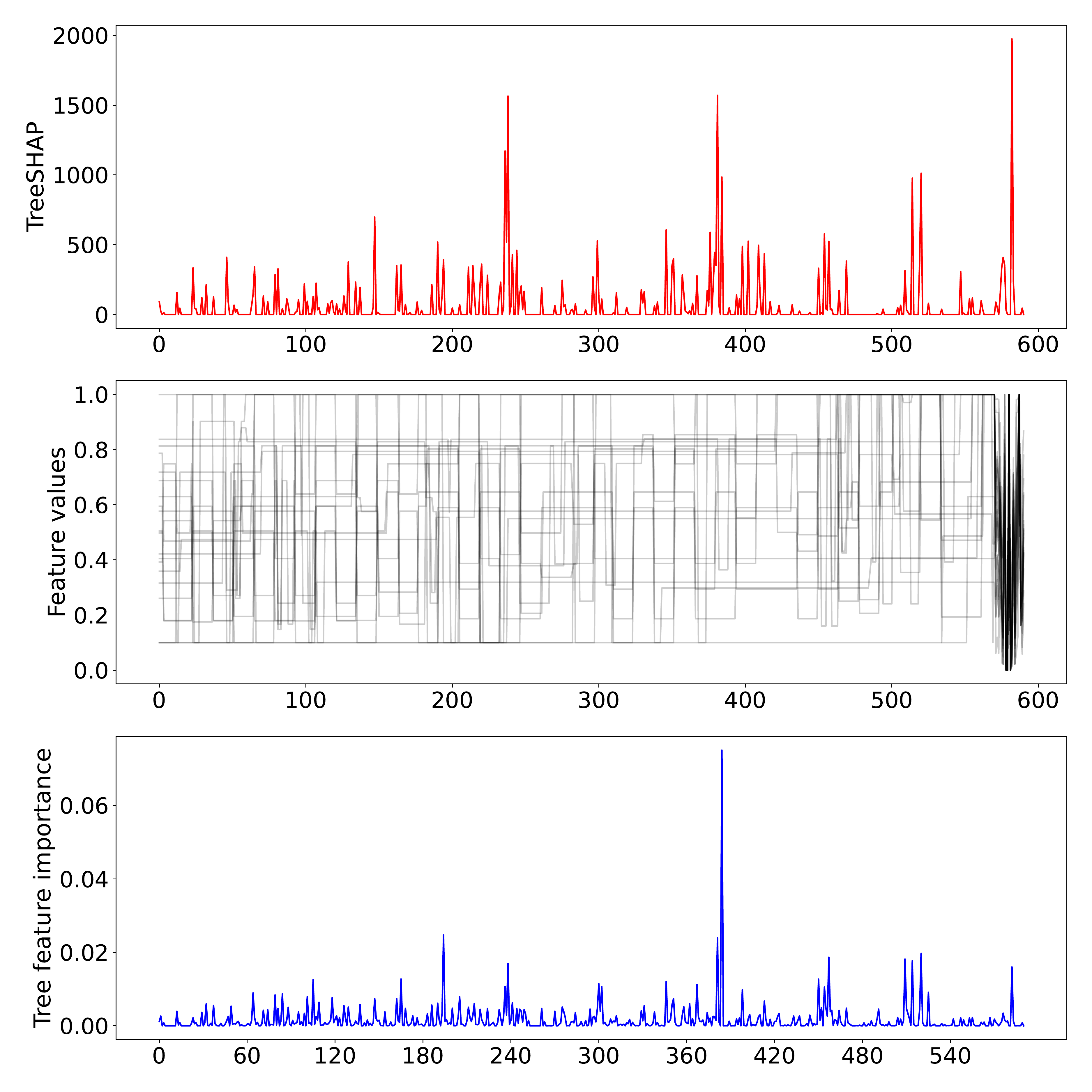}
        \label{fig:pricing_glob_imp_r3}
        }
        \subfloat[Trace dataset ]{\includegraphics[width=0.5\textwidth]{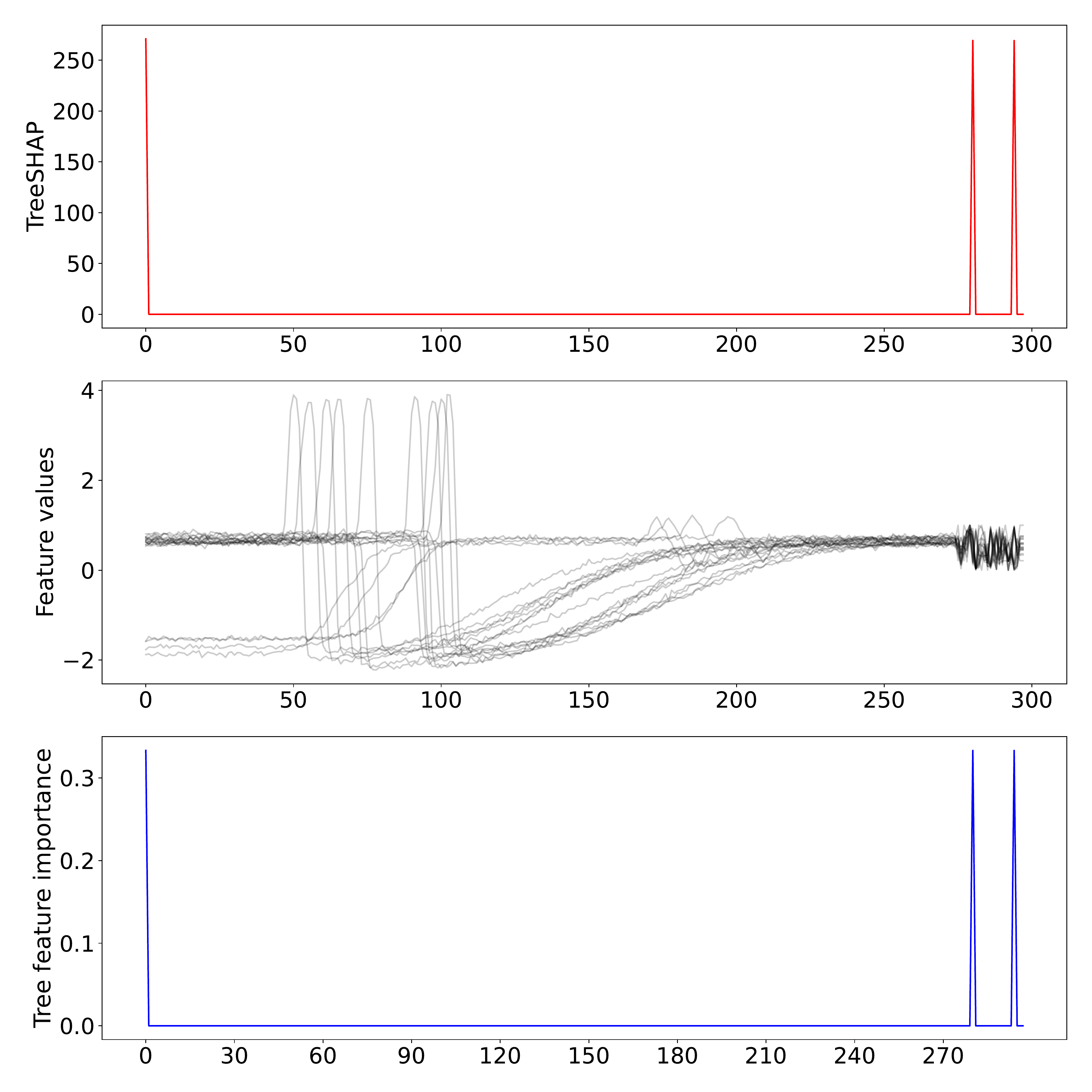}
        \label{fig:trace_glob_imp_r3}
        }\\
        \subfloat[Electricity dataset
        ]{\includegraphics[width=0.5\textwidth]{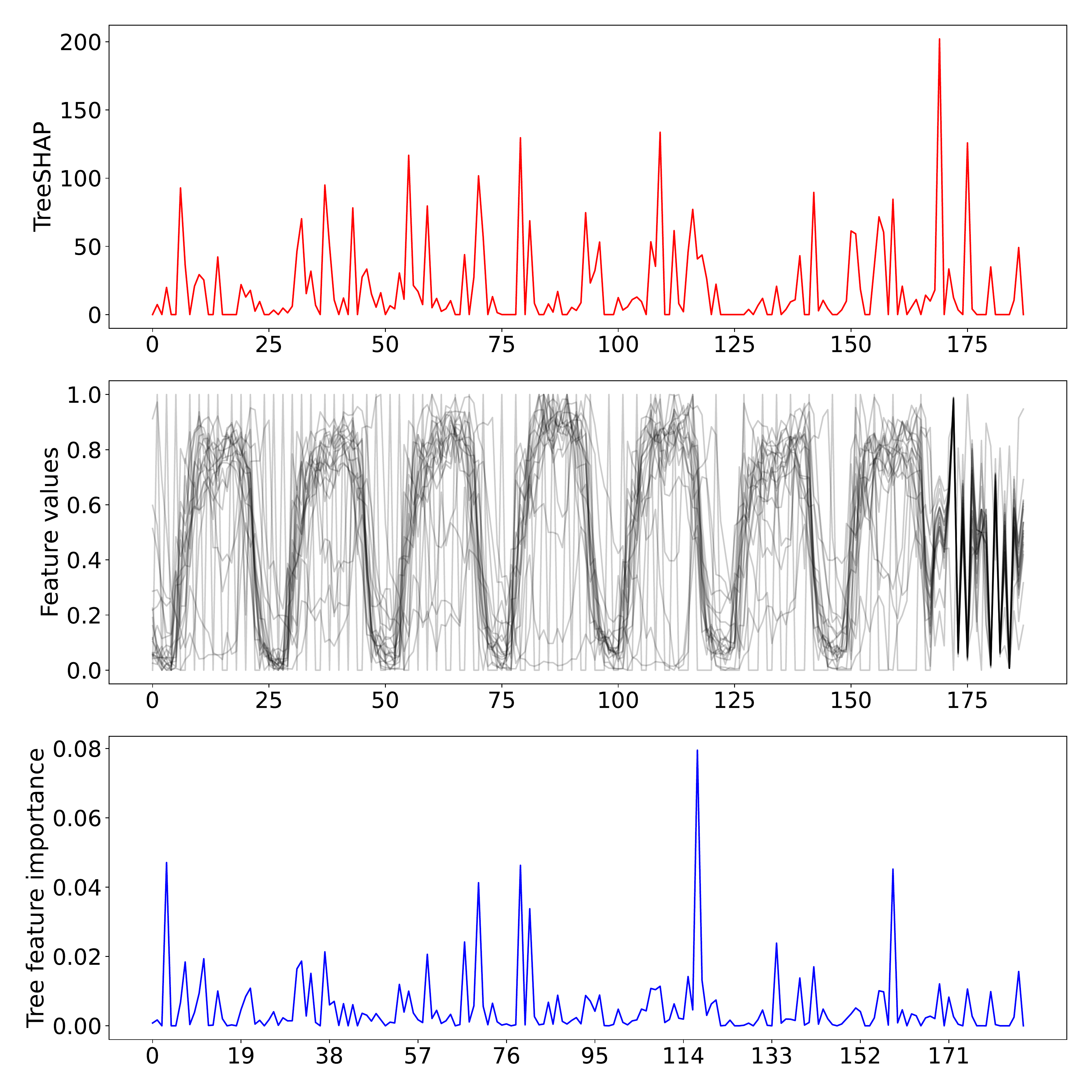}
        \label{fig:electricity_glob_imp_r3}
        }
        \subfloat[Walmart dataset 
        ]{\includegraphics[width=0.5\textwidth]{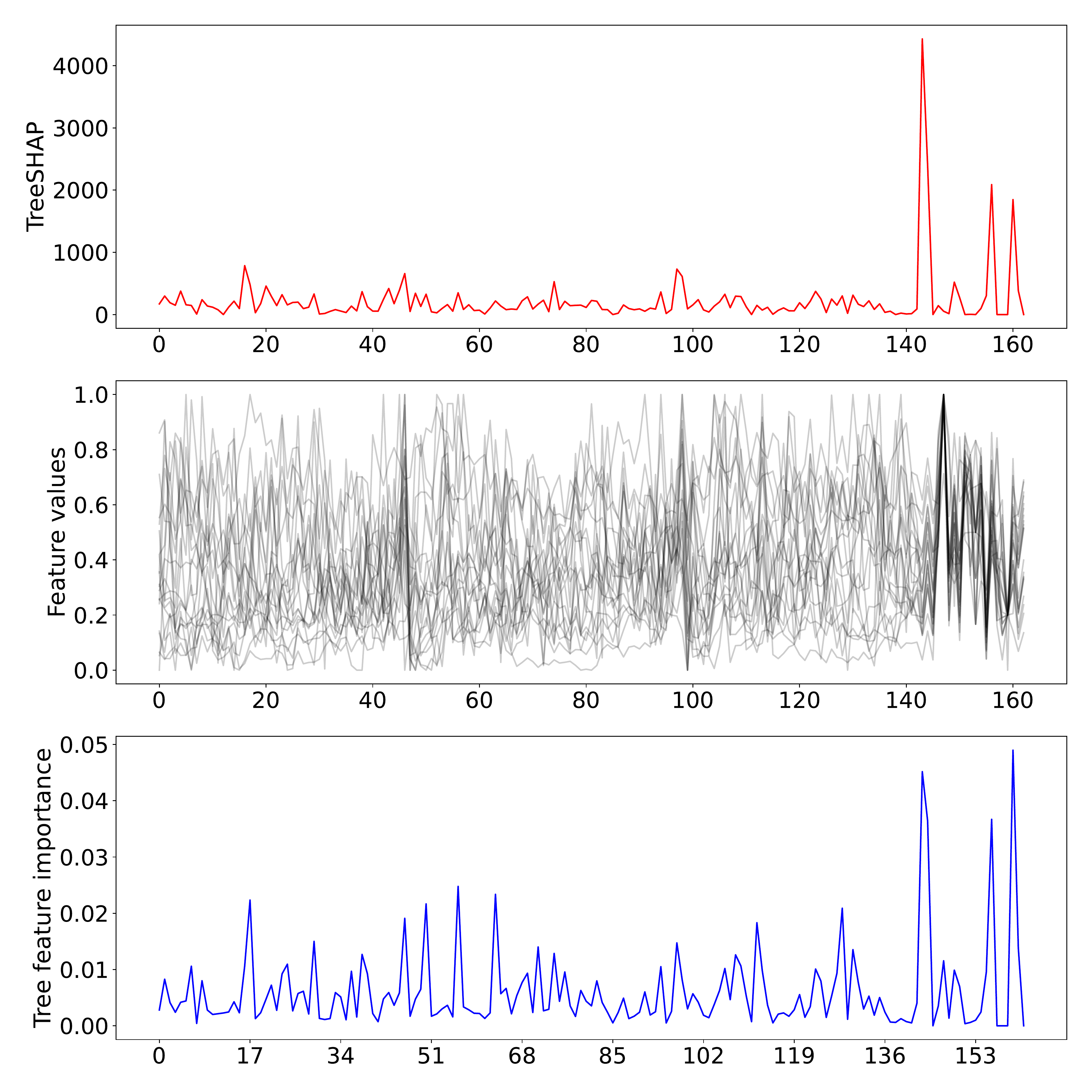}
        \label{fig:walmart_glob_imp_r3}
        }
    \caption{Visualization of global explanations obtained for the XGBoost models using tree feature importance and TreeSHAP interpretability methods. The vertical axis for TreeSHAP and tree feature importance figures show the feature importance, whereas the feature values figure displays the samples' values within the datasets.
    \textbf{($x$-axis is the concatenation of time steps and constructed feature indices, $y$-axis shows the importance attributed to each feature (i.e., timestep or external covariate)).} 
    }
    \label{fig:glob_imp_r3}

\end{figure}

\begin{figure}[hbt!]
    \centering
        \subfloat[Pricing dataset 
        ]{\includegraphics[width=0.5\textwidth]{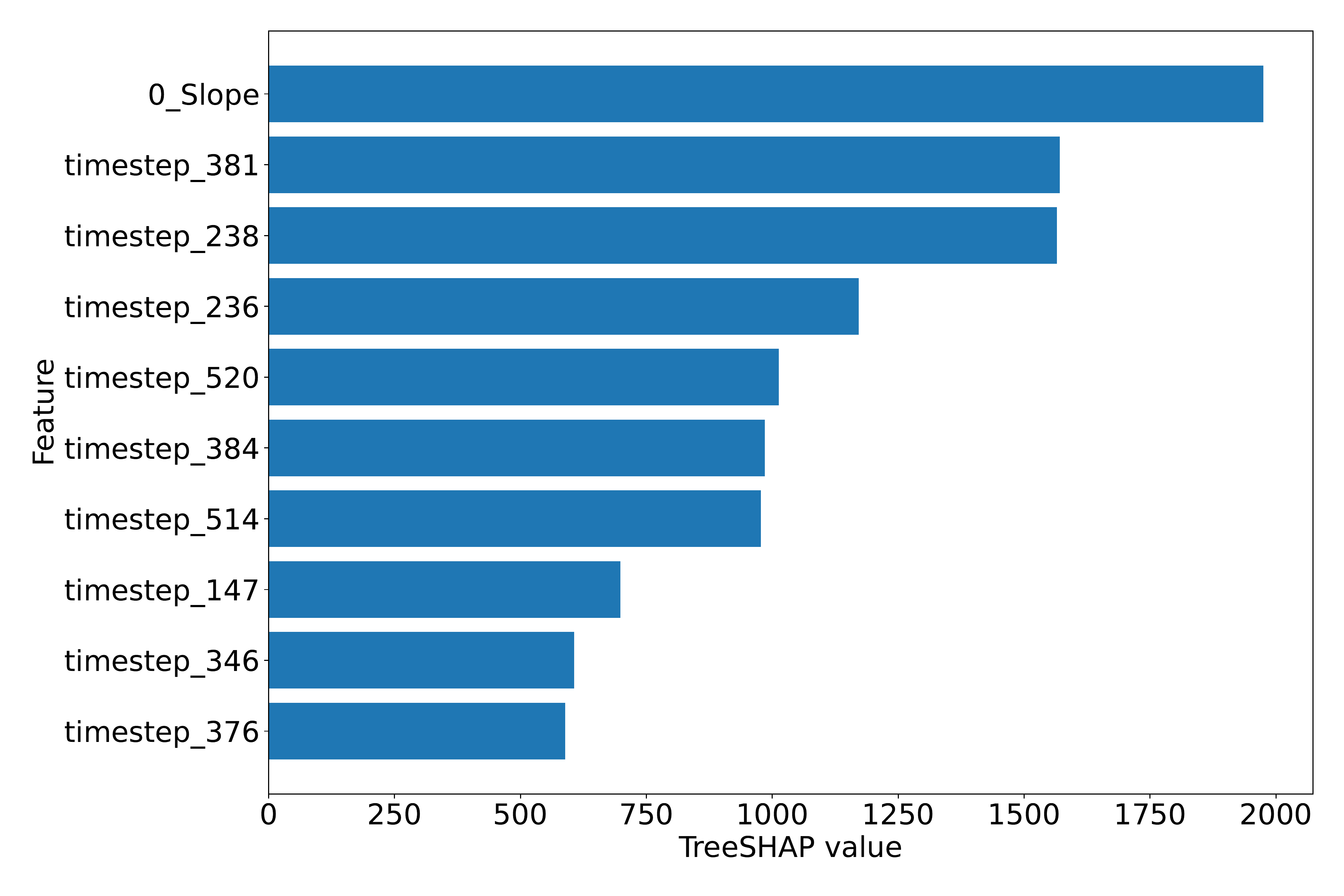}
        \label{fig:pricing_glob_imp_rank_r3}
        }
        \subfloat[Trace dataset ]{\includegraphics[width=0.5\textwidth]{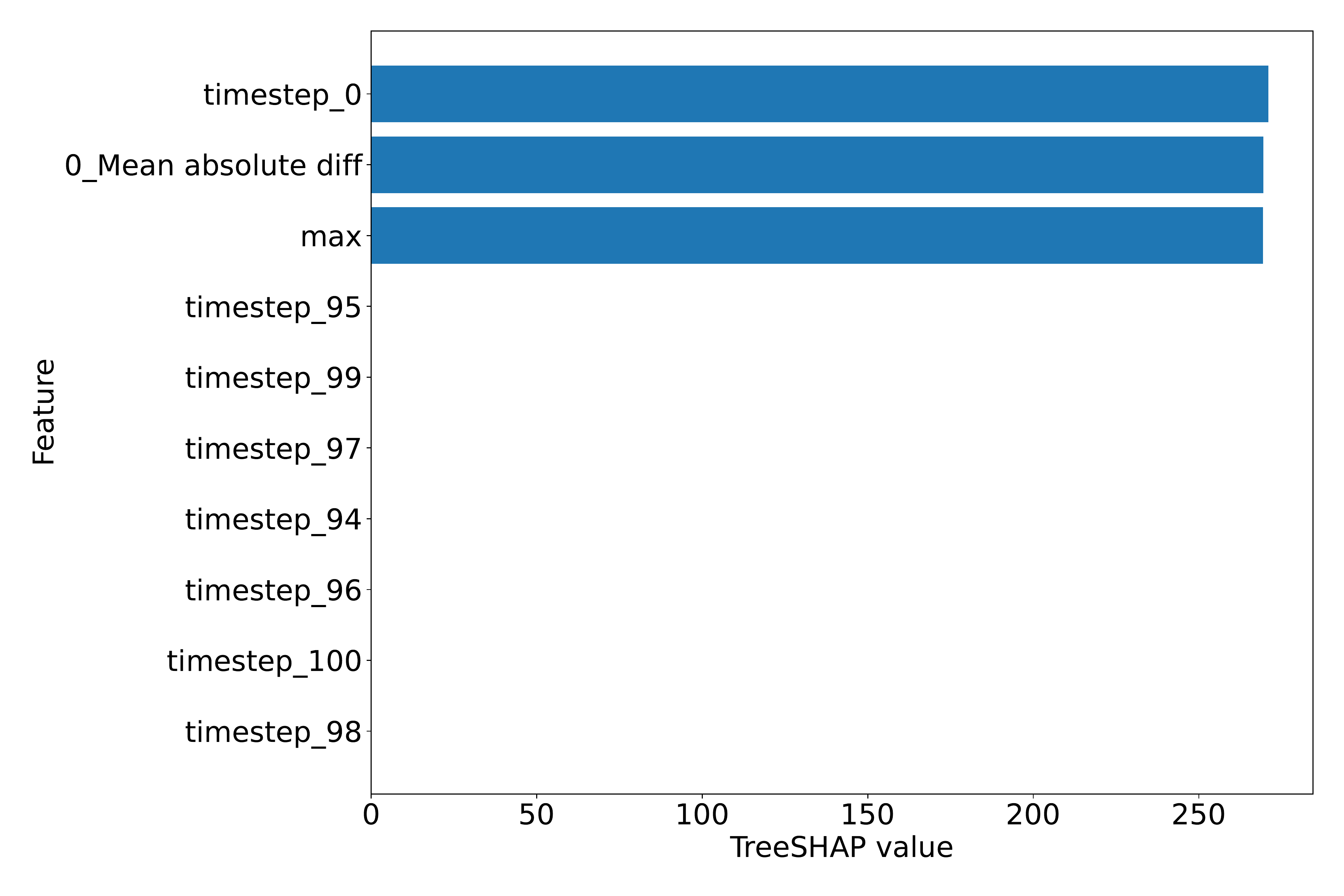}
        \label{fig:trace_glob_imp_rank_r3}
        }\\
        \subfloat[Electricity dataset
        ]{\includegraphics[width=0.5\textwidth]{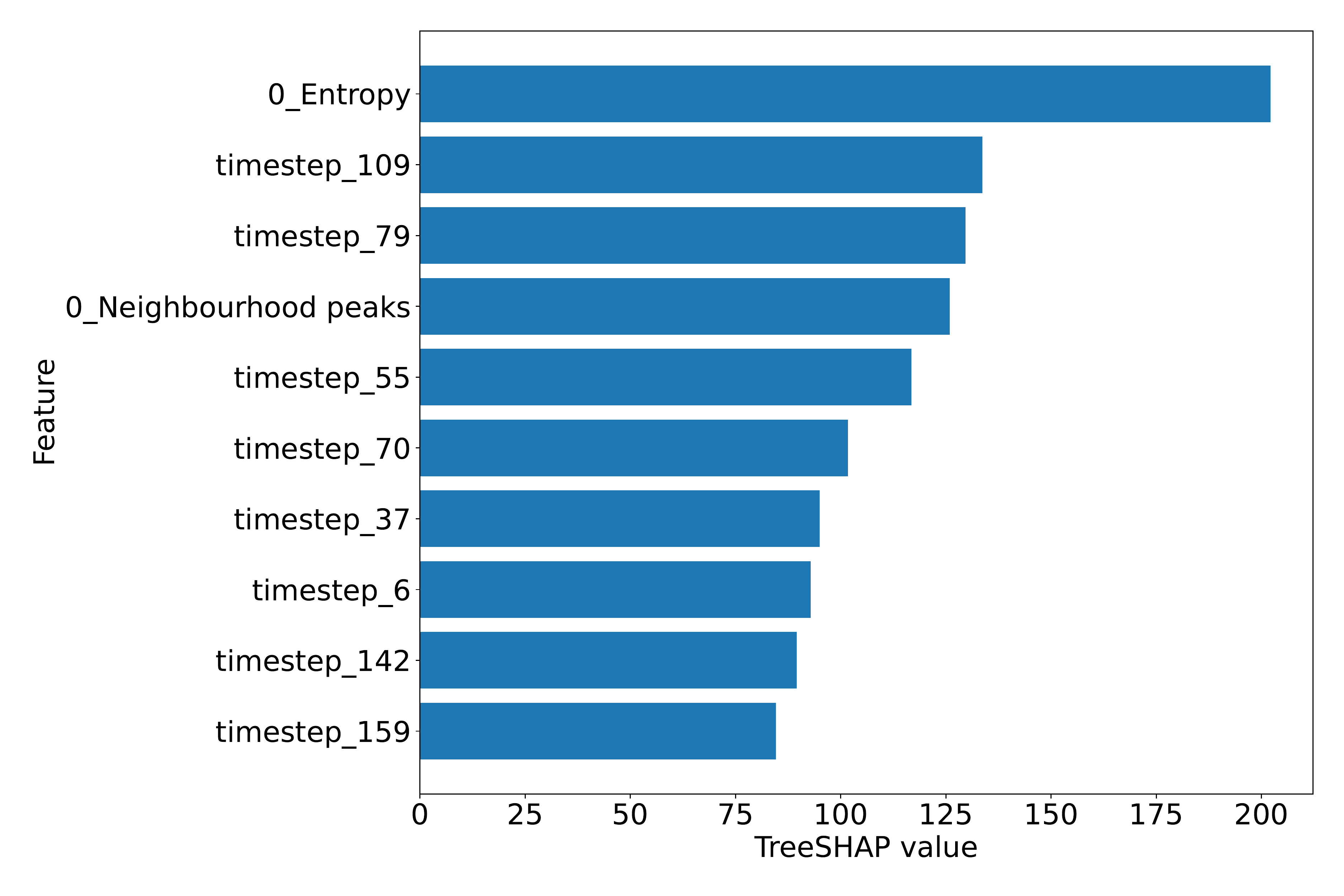}
        \label{fig:electricity_glob_imp_rank_r3}
        }
        \subfloat[Walmart dataset 
        ]{\includegraphics[width=0.5\textwidth]{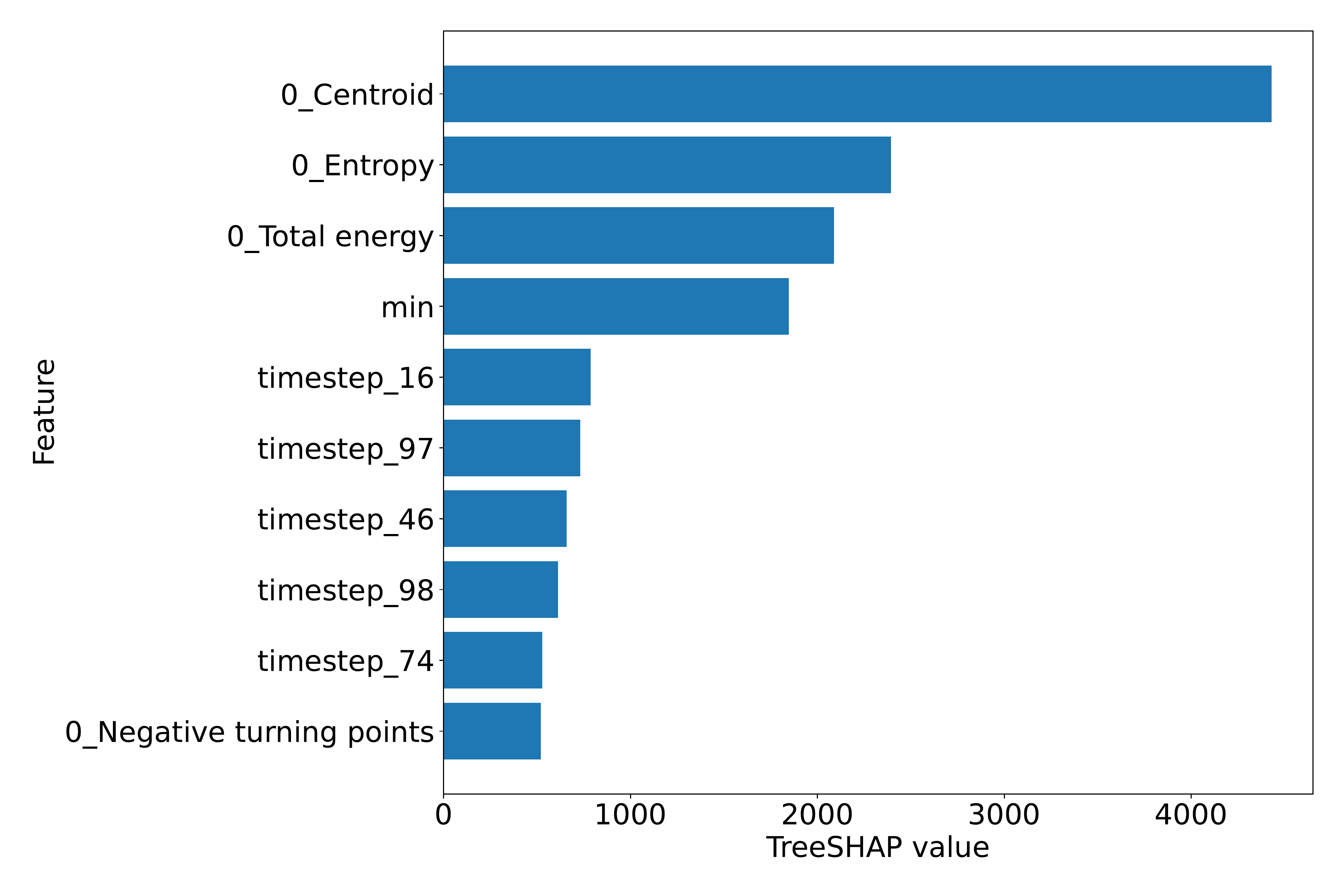}
        \label{fig:walmart_glob_imp_rank_r3}
        }
    \caption{Ranking of the top features in global explanations obtained for the XGBoost models using TreeSHAP interpretability method. }
    \label{fig:glob_imp_rank_r3}

\end{figure}

The global importance of all features, and top features for the XGB model, are shown in Figure~\ref{fig:glob_imp_r3} and Figure~\ref{fig:glob_imp_rank_r3}, respectively. 
We analyze the global importance of the XGB model since it has the highest classification accuracy, and a global explanation can be obtained by calculating the average training loss reduction gained when using a feature for splitting. 
This intrinsic interpretability method is computed using the ``plot\_importances'' function of the \texttt{xgboost} library, and it is referred to as ``tree feature importance'' in the figures. 
We also calculate global explanations using the TreeSHAP method. Since the SHAP computes the contribution of each feature to the prediction for each sample, we take the absolute values of the scores, and then take the average over the samples in the training set to obtain the global importance of the features.  

We find that both explanations match exactly for the trace dataset. We observe that the first timestep of the dataset, mean absolute difference and max are found to be important features for the model predictions. 
The importance of the first timestep can also be seen by comparing the samples from each cluster. 
\tcolA{Among the three classes, samples in Cluster 2 start from negative values, while samples in other clusters start from positive values. }
Thus, the initial timesteps of the dataset are important features for separating the clusters. 
For the Walmart dataset, we find that both interpretability methods find the same three features to be important. These are min, entropy, and centroid.
However, the ranking of the features is different, meaning that both methods do not consistently agree. 
Finally, the tree feature importance method has a more distributed representation of the feature importance, while there is a larger separation between the top features and the others for the TreeSHAP method. 
We observe less consistent results for the pricing and electricity dataset. 
The disagreement is less for the pricing dataset, where some of the top features are identified by both methods, e.g., timesteps 381, 238, and 514. 
Overall, we find that tree feature importance and TreeSHAP explanations do not always agree, especially when the underlying dataset is complex and highly seasonal. 
We note that the differences in the explanations might stem from the interchangeability of the features. 
That is, when features are highly correlated, different assumptions and structures of the interpretability methods can lead to different explanations. 

\subsubsection{Local Feature Importance}\label{sec:local_feat_imp}
\begin{figure}[hbt!]
    \centering
        \subfloat[Pricing dataset - Cluster 18
        ]{\includegraphics[width=0.5\textwidth]{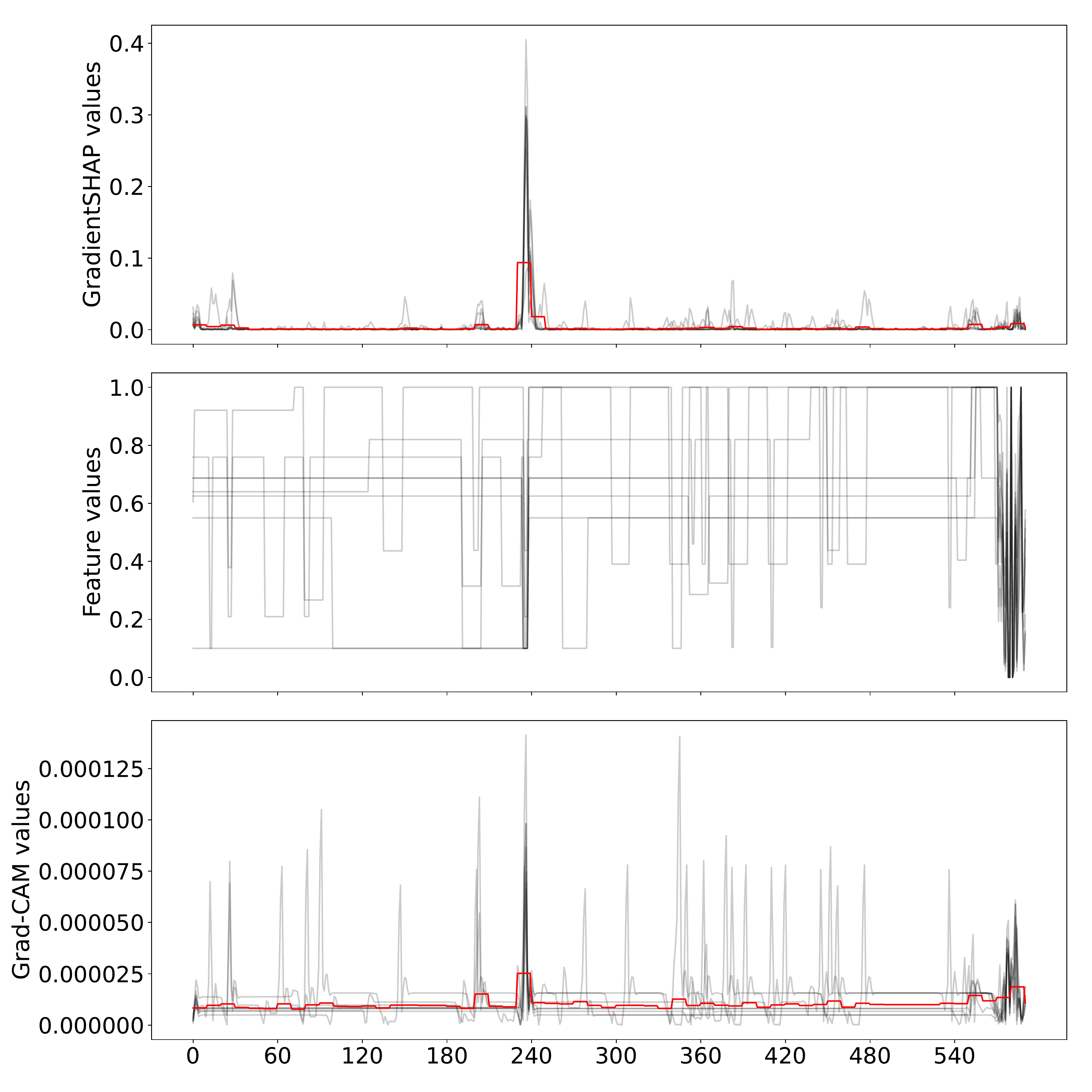}
        \label{fig:pricing_local_imp_18_r3}
        }
        \subfloat[Pricing dataset - Cluster 20
        ]{\includegraphics[width=0.5\textwidth]{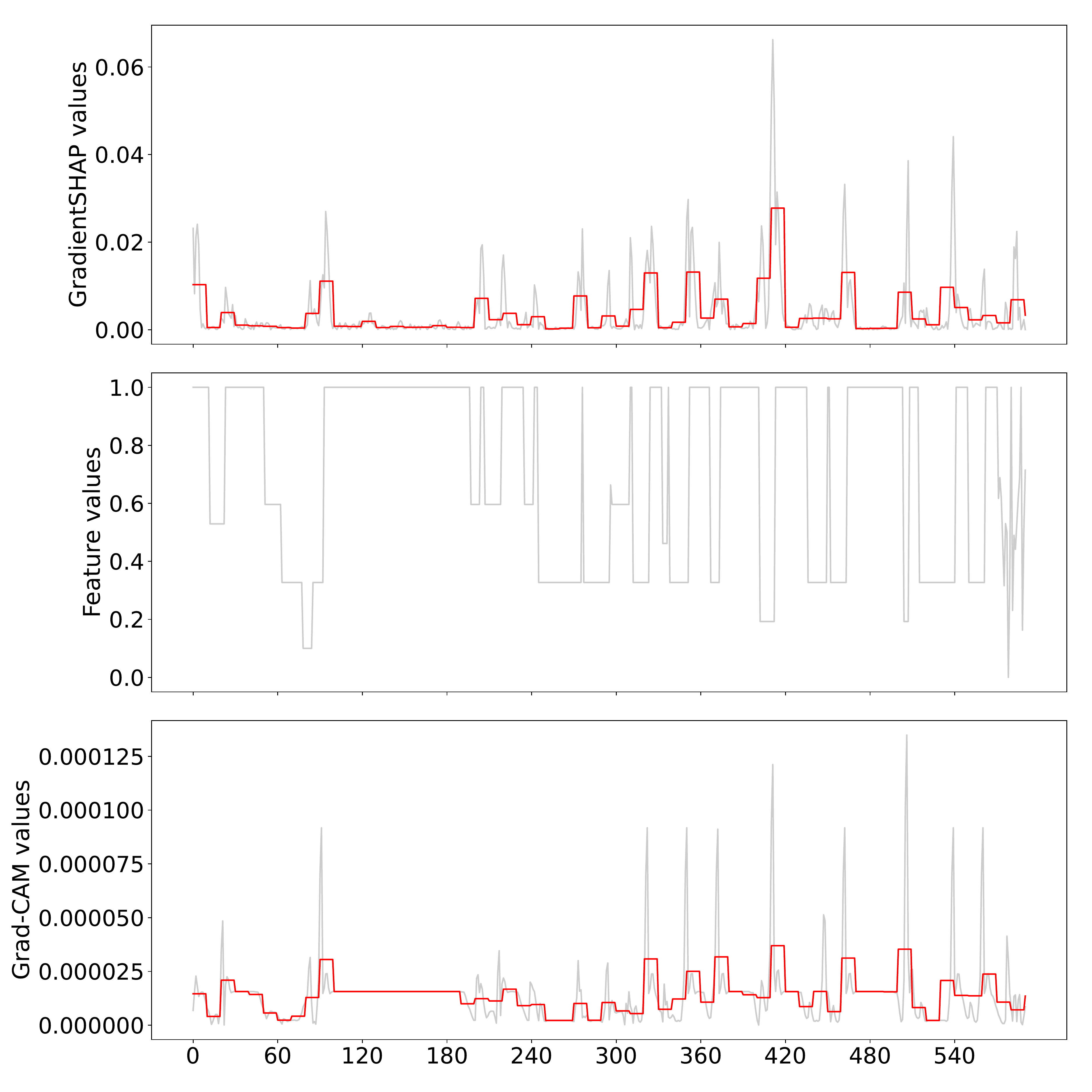}
        \label{fig:pricing_local_imp_20_r3}
        }\\
    \caption{Visualization of cluster instances and explanations for the pricing dataset.
    ($x$-axis is the concatenation of time steps and constructed feature indices).
    }
    \label{fig:visualize_local_imp_pricing_r3}

\end{figure}

\begin{figure}[hbt!]
    \centering
        \subfloat[Trace dataset - Cluster 0 ]{\includegraphics[width=0.5\textwidth]{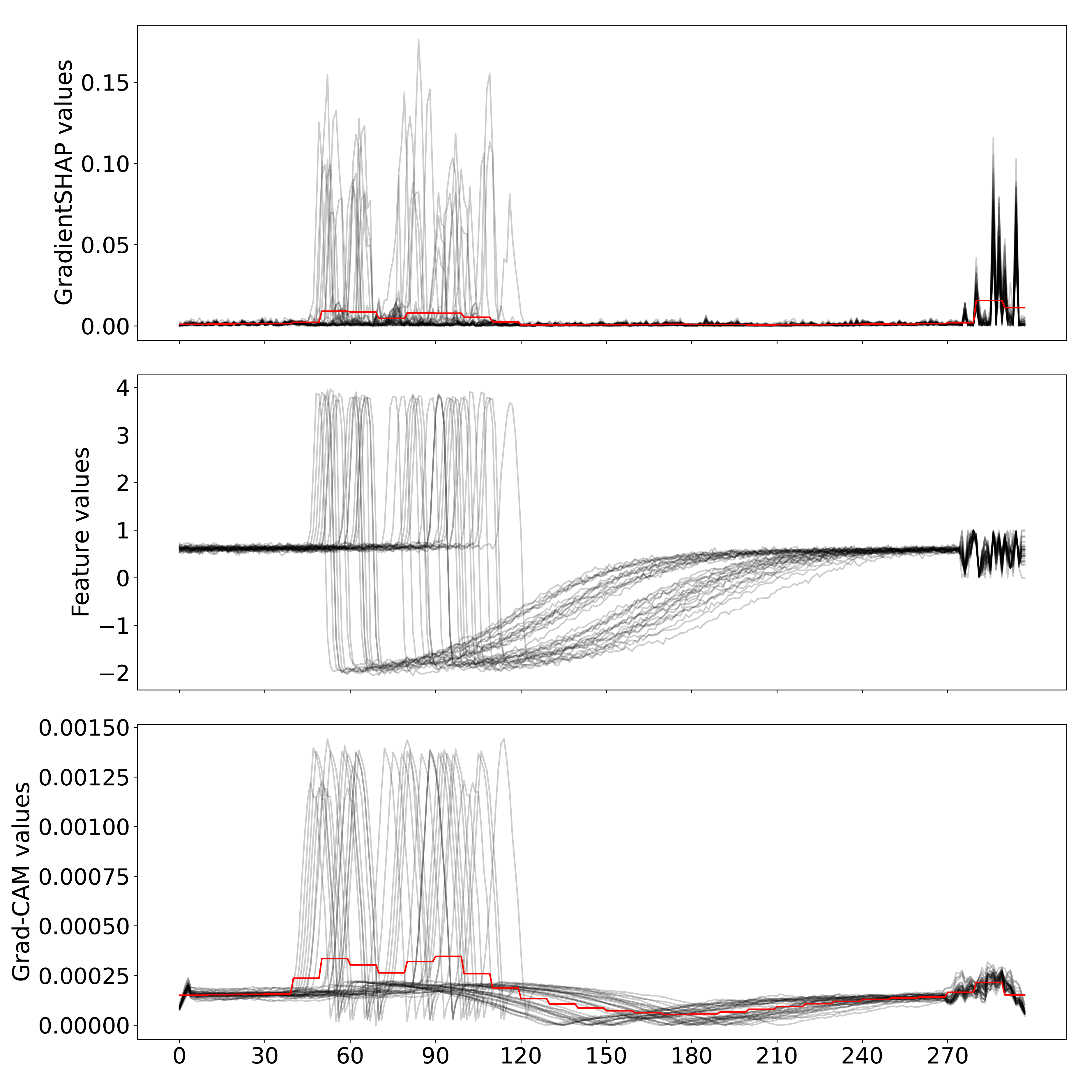}
        \label{fig:trace_pred_vis_0_r3}
        }
        \subfloat[Trace dataset - Cluster 2 ]{\includegraphics[width=0.5\textwidth]{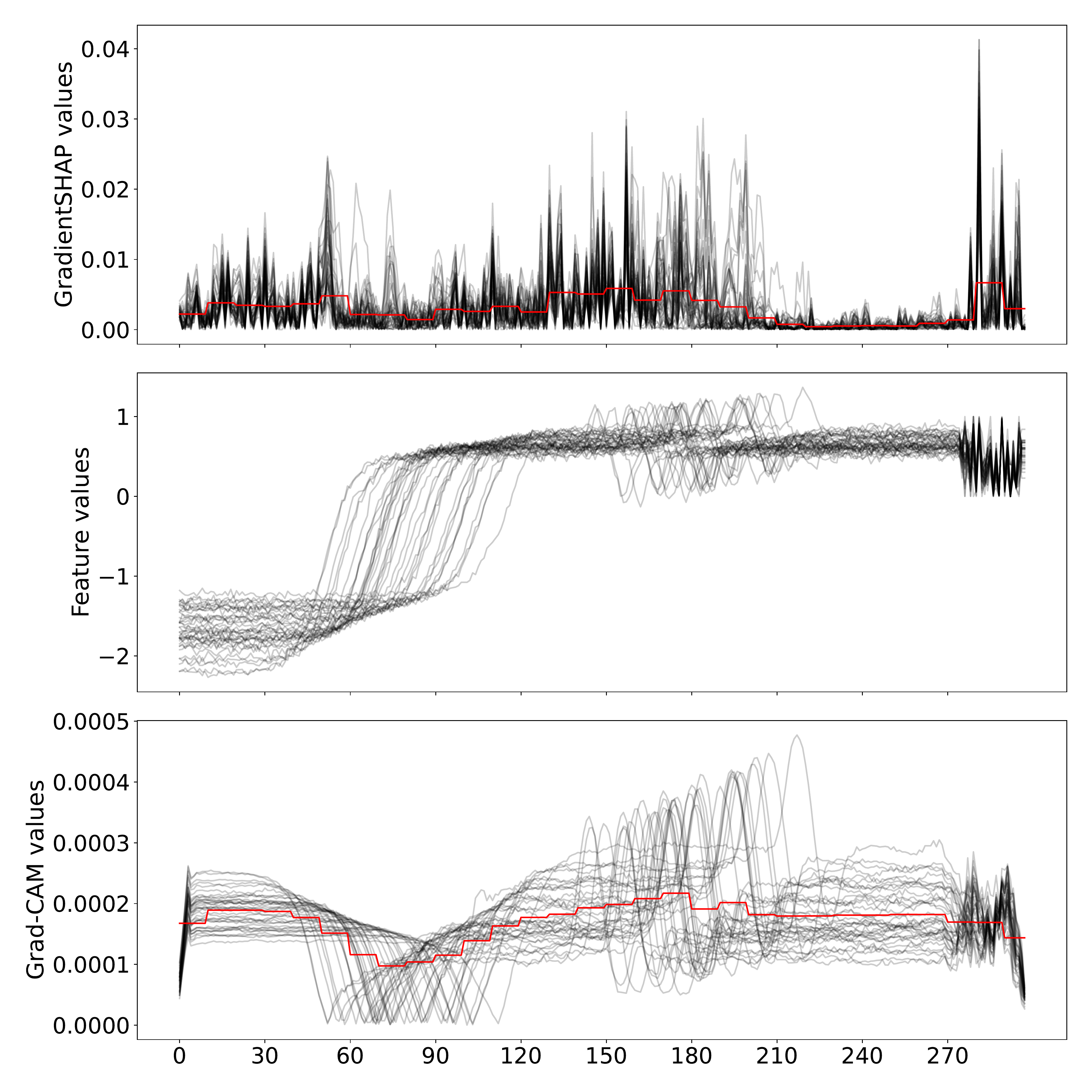}
        \label{fig:trace_pred_vis_2_r3}
        }\\
        
    \caption{Visualization of cluster instances and explanations for the trace dataset.
    ($x$-axis is the concatenation of time steps and constructed feature indices).
    }
    \label{fig:visualize_local_imp_trace_r3}

\end{figure}

\begin{figure}[hbt!]
    \centering
        \subfloat[Electricity dataset - Cluster 4
        ]{\includegraphics[width=0.5\textwidth]{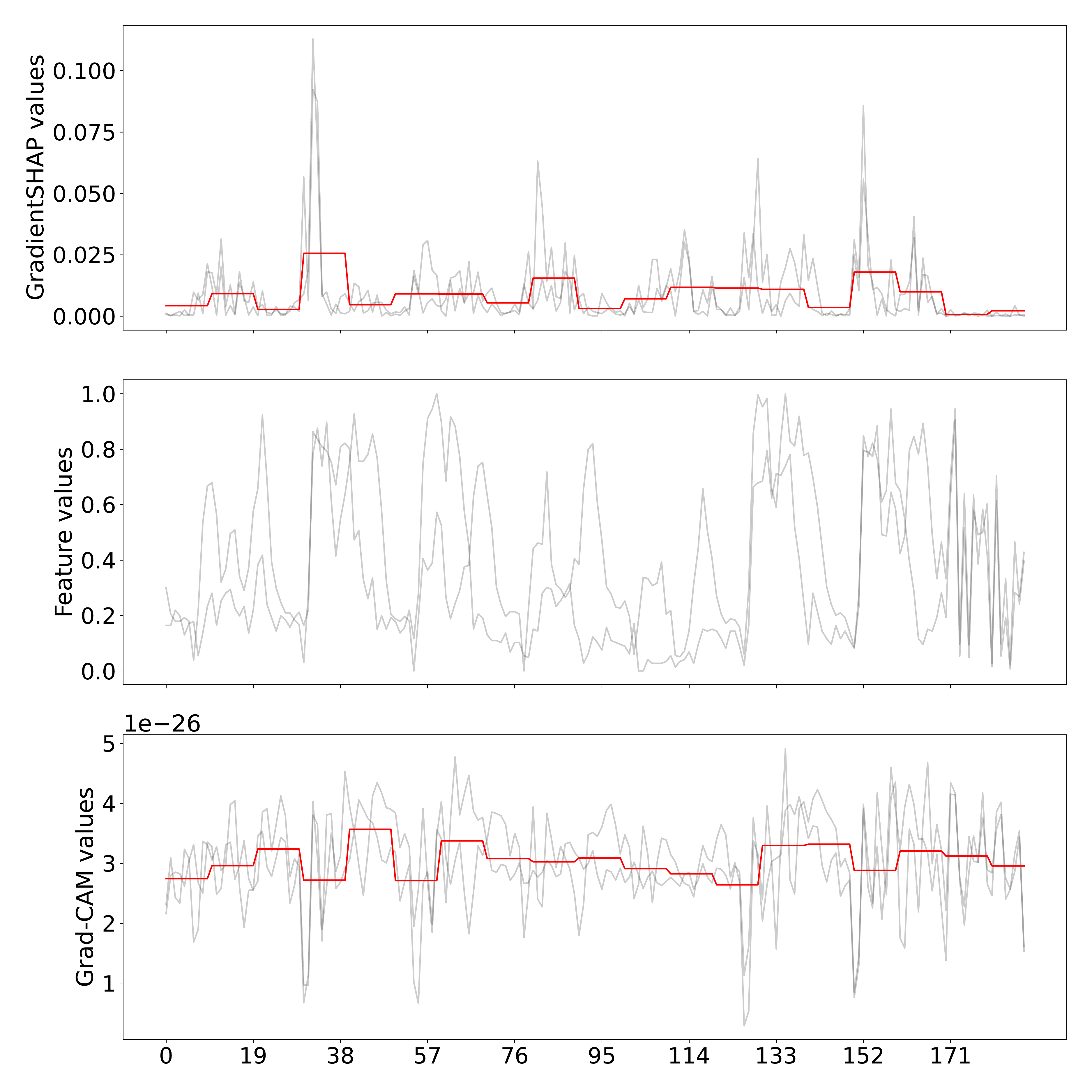}
        \label{fig:electricity_pred_vis_4_r3}
        }
        \subfloat[Electricity dataset - Cluster 12
        ]{\includegraphics[width=0.5\textwidth]{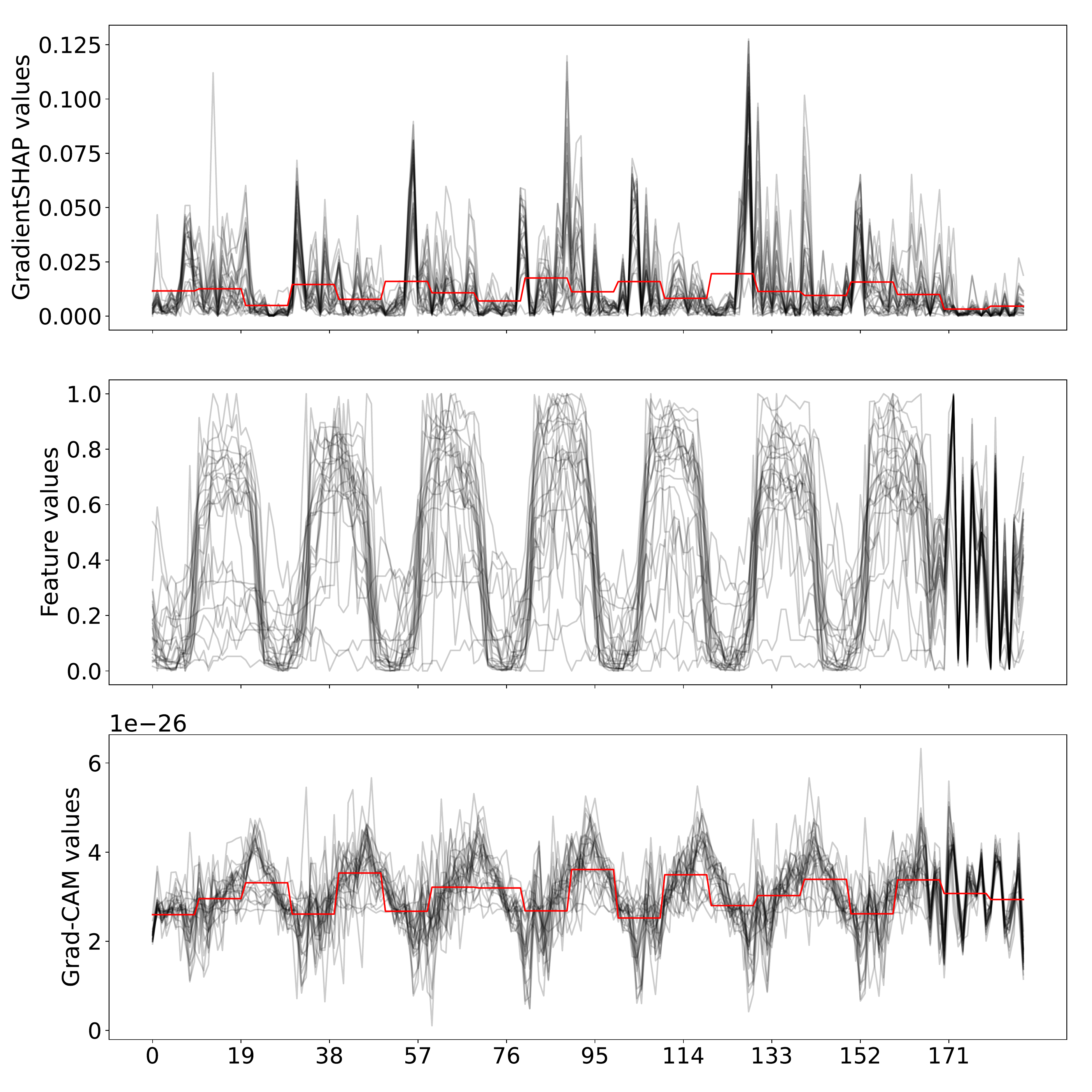}
        \label{fig:electricity_pred_vis_12_r3}
        }\\
    \caption{Visualization of cluster instances and explanations for the electricity dataset.
    ($x$-axis is the concatenation of time steps and constructed feature indices).
    }
    \label{fig:visualize_local_imp_electricity_r3}

\end{figure}

\begin{figure}[hbt!]
    \centering
        \subfloat[Walmart dataset - Cluster 1
        ]{\includegraphics[width=0.5\textwidth]{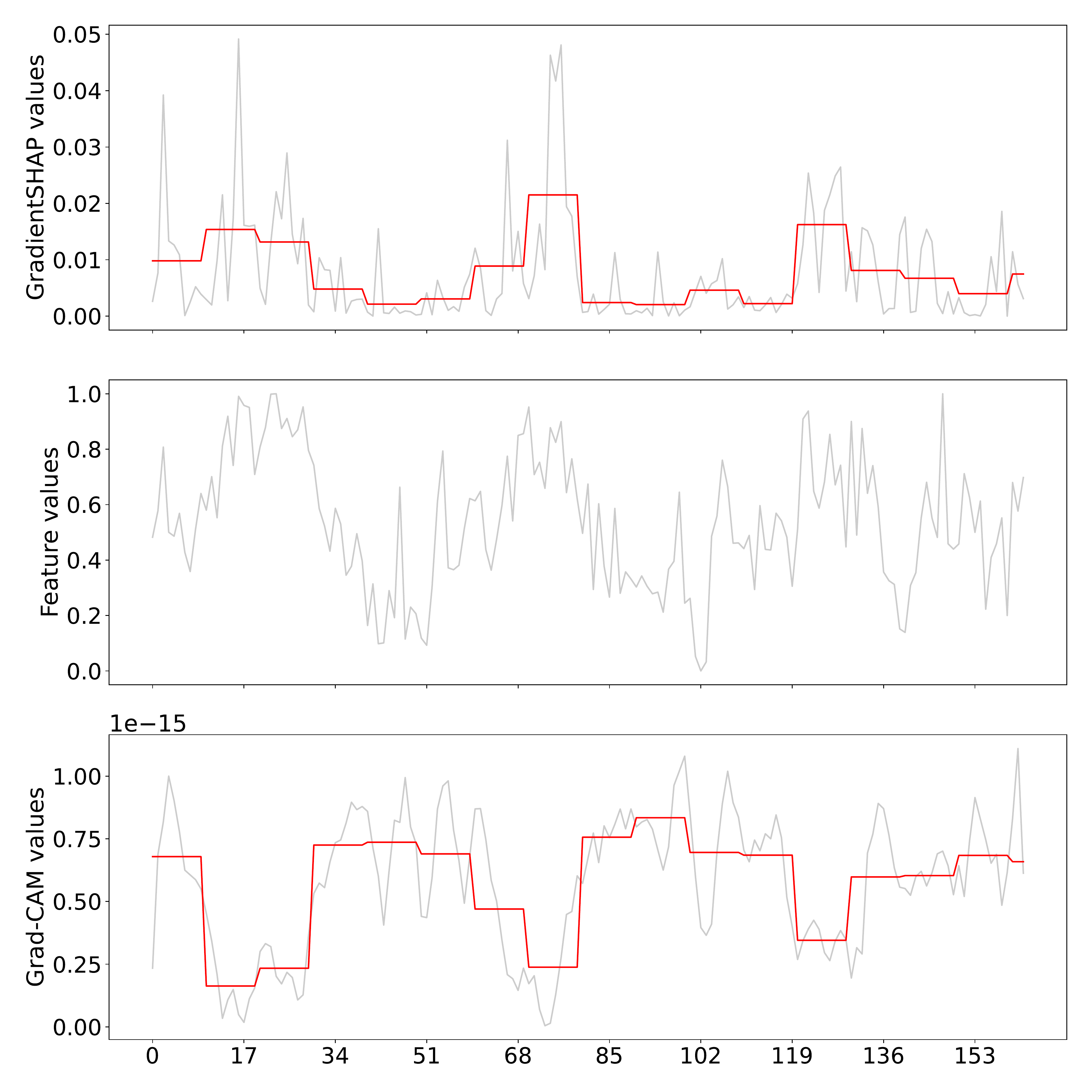}
        \label{fig:walmart_pred_vis_1_r3}
        }
        \subfloat[Walmart dataset - Cluster 11
        ]{\includegraphics[width=0.5\textwidth]{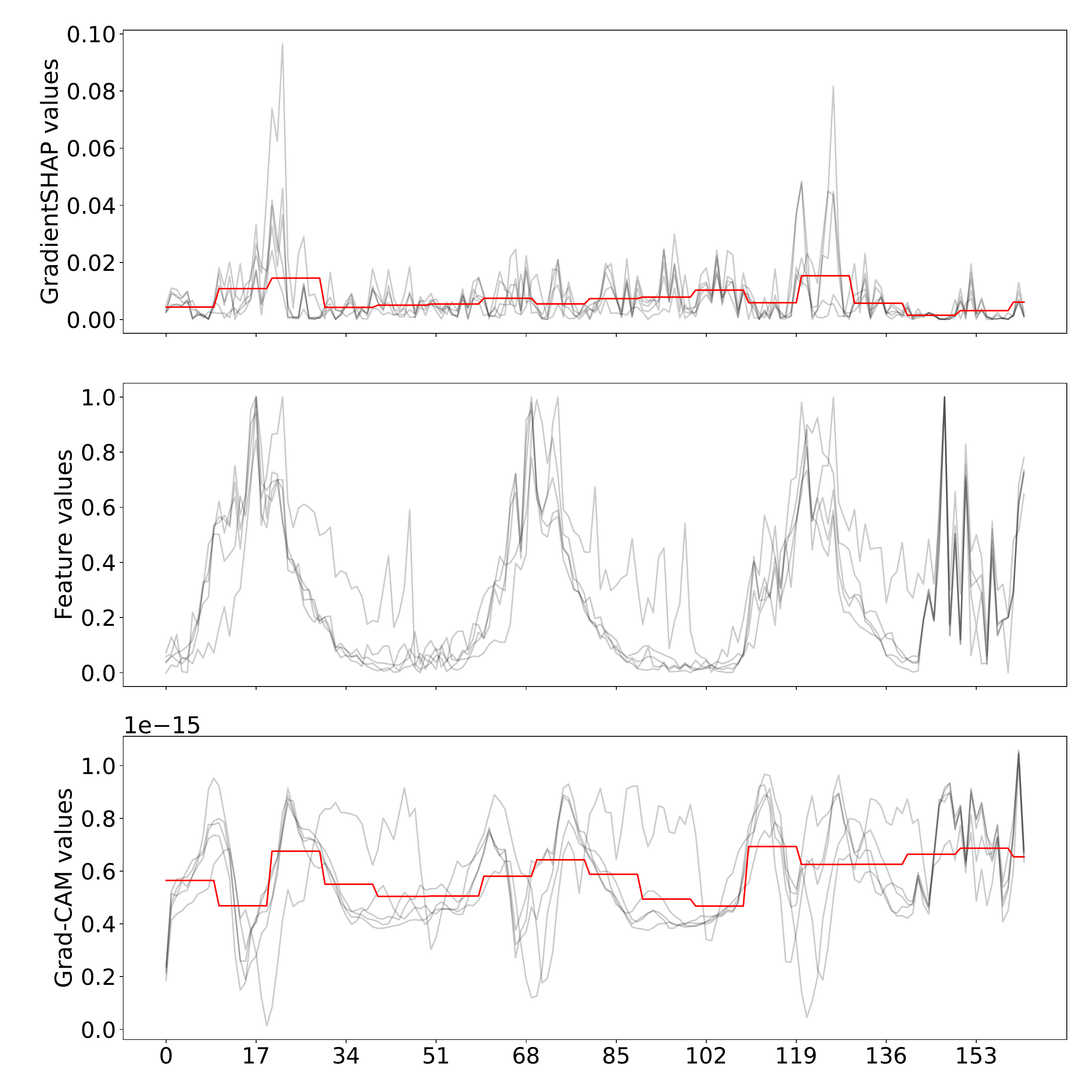}
        \label{fig:walmart_pred_vis_11_r3}
        }\\
        
    \caption{Visualization of cluster instances and explanations for the Walmart dataset.
    ($x$-axis is the concatenation of time steps and constructed feature indices).
    }
    \label{fig:visualize_local_imp_walmart_r3}

\end{figure}

We use the FCN model for analysis of the local feature importance, as both Grad-CAM and GradientSHAP methods can be used to interpret the FCN models' predictions. 
With both interpretability methods, we compute the feature importance for all the samples in two representative clusters. 
The results for all 4 datasets are presented in Figure~\ref{fig:visualize_local_imp_pricing_r3} (pricing), Figure~\ref{fig:visualize_local_imp_trace_r3} (trace), Figure~\ref{fig:visualize_local_imp_electricity_r3} (electricity), and Figure~\ref{fig:visualize_local_imp_walmart_r3} (Walmart). 
The first and last rows of each figure show the computed feature importance for instances of the clusters in gray, and the average feature importances for a selected cluster in red. 
To provide aggregate importance of the features, the feature importances are averaged over time windows of five timesteps (e.g., 0-5, 5-10, 10-15). 

In the pricing dataset, we see that both methods generally agree on the important features. 
Because the MPBD metric is used for training the clustering model, we know that the price change points should be the important factors for clustering the instances. 
On the other hand, time windows where the price is stable are less impactful, as it has a low impact on MPBD. 
As expected, from Figure~\ref{fig:pricing_local_imp_18_r3}, and Figure~\ref{fig:pricing_local_imp_20_r3}, we see the explanations assigning more importance to these price change points. 
Interestingly, for Cluster 18, the Grad-CAM method assigns importance to many of the price change points, whereas GradientSHAP attends very high importance to a short time window (235-240). 
However, for Cluster 20, we find that both methods attend to similar time windows, with a high agreement on which points are important for the prediction of the clusters. 

In the trace dataset, both interpretability methods assign higher importance to the extracted features. 
The explanations are highly correlated for Cluster 0, compared to Cluster 2, in which two time ranges (30-120 and 120-210) are likely important for the prediction. 
Surprisingly, the Grad-CAM method finds that the first part of the instances (30-120) is less important than the importance of a random point in the instance. 
This is likely incorrect since the data points where there is no change (e.g., 0-30, 210-260) are stable for all the clusters in the dataset. 

In electricity and Walmart datasets, we find that there is a high disagreement between GradientSHAP and Grad-CAM, and the GradientSHAP method is more consistent for the parts it finds significant. 
Both in Figure~\ref{fig:electricity_pred_vis_4_r3} and Figure~\ref{fig:walmart_pred_vis_11_r3}, the Grad-CAM method does not clearly identify any of the features as important, whereas assigns much larger importance to a small set of data points. 
Furthermore, in Figure~\ref{fig:electricity_pred_vis_12_r3} and Figure~\ref{fig:walmart_pred_vis_11_r3}, GradientSHAP assigns larger importance to certain parts of the seasonal cycles. 
In contrast, Grad-CAM assigns almost equal importance to similar sections of the seasonal cycles. 
Overall, we find that there is a high disagreement between the two interpretability methods, with GradientSHAP producing more logical and coherent explanations of clusters, and individual instances.

\section{Conclusion}\label{sec:conclusion_r3}
In this work, we study the use of global and local interpretability methods for explaining time series clustering models. 
Our approach involves training time series classification models to predict the cluster labels. 
Afterwards, local interpretability methods are used to interpret the classification models, and the explanations are used to explain the clustering models. 
To evaluate the effectiveness of the proposed methodology, we perform numerical experiments on multiple time series datasets, clustering models, and classification models. 
We then analyze the results and compare different model interpretability methods along with the generated global and local explanations. 
We motivate the practical usage of the proposed methodology through a product pricing problem. 
Based on our detailed numerical experiments, we observe that the interpretability methods can more accurately identify salient data points, particularly when the dataset is simpler, and has low seasonality. 
We also find that, for the local explanations, the SHAP variants perform better than the Grad-CAM method at correctly identifying the salient data points as important. 
An interesting area to further explore is the use of carefully constructed features to train the classification models. 
Using domain knowledge, it is possible to construct highly interpretable and predictive features, which can then be used to train the classification models without using the raw data points (i.e., values at time steps) as the inputs. 
Subsequently, a clustering model's behavior can be explained through the variable importance analysis of the extracted features. 
A drawback of the proposed methodology is that it requires retraining of the classification model whenever new data arrives. 
Online machine learning techniques can be incorporated to reduce the amount of computation required to learn from the newly arriving data.

\hfill \break
{\bf Acknowledgement.} The authors would like to thank Getir\footnote{https://getir.com/} for supporting this project.


\section*{Data availability statement}
All the datasets except for pricing data are publicly available, and can be obtained using the cited sources.

\section*{Disclosure statement}
No potential conflict of interest was reported by the authors.

\bibliographystyle{spbasic} 
\bibliography{refs}

\begin{thebibliography}{25}
\providecommand{\natexlab}[1]{#1}
\providecommand{\url}[1]{{#1}}
\providecommand{\urlprefix}{URL }
\expandafter\ifx\csname urlstyle\endcsname\relax
  \providecommand{\doi}[1]{DOI~\discretionary{}{}{}#1}\else
  \providecommand{\doi}{DOI~\discretionary{}{}{}\begingroup
  \urlstyle{rm}\Url}\fi
\providecommand{\eprint}[2][]{\url{#2}}

\bibitem[{Adadi and Berrada(2018)}]{adadi2018peeking}
Adadi A, Berrada M (2018) Peeking {{Inside}} the {{Black}}-{{Box}}: {{A
  Survey}} on {{Explainable Artificial Intelligence}} ({{XAI}}). IEEE Access
  6:52138--52160

\bibitem[{Alvarez et~al.(2010)Alvarez, Troncoso, Riquelme, and
  Ruiz}]{alvarez2010energy}
Alvarez FM, Troncoso A, Riquelme JC, Ruiz JSA (2010) Energy time series
  forecasting based on pattern sequence similarity. IEEE Transactions on
  Knowledge and Data Engineering 23(8):1230--1243

\bibitem[{Barandas et~al.(2020)Barandas, Folgado, Fernandes, Santos, Abreu,
  Bota, Liu, Schultz, and Gamboa}]{barandas2020tsfel}
Barandas M, Folgado D, Fernandes L, Santos S, Abreu M, Bota P, Liu H, Schultz
  T, Gamboa H (2020) Tsfel: Time series feature extraction library. SoftwareX
  11:100456

\bibitem[{Bertsimas et~al.(2021)Bertsimas, Orfanoudaki, and
  Wiberg}]{bertsimas2021interpretable}
Bertsimas D, Orfanoudaki A, Wiberg H (2021) Interpretable clustering: an
  optimization approach. Machine Learning 110(1):89--138

\bibitem[{Bojer and Meldgaard(2021)}]{bojer2021kaggle}
Bojer CS, Meldgaard JP (2021) Kaggle forecasting competitions: An overlooked
  learning opportunity. International Journal of Forecasting 37(2):587--603

\bibitem[{Bozanta et~al.(2022)Bozanta, Berry, Cevik, Bulut, Yigit, Gonen, and
  Ba{\c{s}}ar}]{bozanta2022time}
Bozanta A, Berry S, Cevik M, Bulut B, Yigit D, Gonen FF, Ba{\c{s}}ar A (2022)
  Time series clustering for grouping products based on price and sales
  patterns. arXiv preprint arXiv:220408334

\bibitem[{Cannone et~al.(2020)Cannone, Baralis, and
  Pastor}]{cannone2020explainable}
Cannone M, Baralis E, Pastor DE (2020) Explainable ai for clustering
  algorithms. Master's thesis, Politecnico Di Torino

\bibitem[{Caruana(2017)}]{caruana2017intelligible}
Caruana R (2017) Intelligible machine learning for critical applications such
  as health care. In: 2017 AAAS Annual Meeting (February 16-20, 2017), AAAS

\bibitem[{Dua and Graff(2017)}]{Dua:2019}
Dua D, Graff C (2017) {UCI} machine learning repository.
  \urlprefix\url{http://archive.ics.uci.edu/ml}

\bibitem[{Fauvel et~al.(2021)Fauvel, Lin, Masson, Fromont, and
  Termier}]{fauvel2021xcm}
Fauvel K, Lin T, Masson V, Fromont {\'E}, Termier A (2021) Xcm: An explainable
  convolutional neural network for multivariate time series classification.
  Mathematics 9(23):3137

\bibitem[{Ismail et~al.(2020)Ismail, Gunady, Corrada~Bravo, and
  Feizi}]{ismail2020benchmarking}
Ismail AA, Gunady M, Corrada~Bravo H, Feizi S (2020) Benchmarking deep learning
  interpretability in time series predictions. Advances in neural information
  processing systems 33:6441--6452

\bibitem[{Ismail~Fawaz et~al.(2019)Ismail~Fawaz, Forestier, Weber, Idoumghar,
  and Muller}]{ismail2019deep}
Ismail~Fawaz H, Forestier G, Weber J, Idoumghar L, Muller PA (2019) Deep
  learning for time series classification: a review. Data mining and knowledge
  discovery 33(4):917--963

\bibitem[{Kodinariya and Makwana(2013)}]{kodinariya2013review}
Kodinariya TM, Makwana PR (2013) Review on determining number of cluster in
  k-means clustering. International Journal 1(6):90--95

\bibitem[{Lundberg and Lee(2017)}]{lundberg2017unified}
Lundberg SM, Lee SI (2017) A unified approach to interpreting model
  predictions. Advances in Neural Information Processing Systems
  2017-Decem(Section 2):4766--4775

\bibitem[{Lundberg et~al.(2020)Lundberg, Erion, Chen, DeGrave, Prutkin, Nair,
  Katz, Himmelfarb, Bansal, and Lee}]{lundberg2020local}
Lundberg SM, Erion G, Chen H, DeGrave A, Prutkin JM, Nair B, Katz R, Himmelfarb
  J, Bansal N, Lee SI (2020) From local explanations to global understanding
  with explainable {{AI}} for trees. Nature Machine Intelligence 2(1):56--67

\bibitem[{Marcinkevi{\v{c}}s and
  Vogt(2020)}]{marcinkevivcs2020interpretability}
Marcinkevi{\v{c}}s R, Vogt JE (2020) Interpretability and explainability: A
  machine learning zoo mini-tour. arXiv preprint arXiv:201201805

\bibitem[{Meesrikamolkul et~al.(2012)Meesrikamolkul, Niennattrakul, and
  Ratanamahatana}]{meesrikamolkul2012shape}
Meesrikamolkul W, Niennattrakul V, Ratanamahatana CA (2012) Shape-based
  clustering for time series data. In: Pacific-Asia Conference on Knowledge
  Discovery and Data Mining, Springer, pp 530--541

\bibitem[{Ozyegen et~al.(2022)Ozyegen, Ilic, and Cevik}]{ozyegen2022evaluation}
Ozyegen O, Ilic I, Cevik M (2022) Evaluation of interpretability methods for
  multivariate time series forecasting. Applied Intelligence 52(5):4727--4743

\bibitem[{R{\"a}s{\"a}nen and Kolehmainen(2009)}]{rasanen2009feature}
R{\"a}s{\"a}nen T, Kolehmainen M (2009) Feature-based clustering for
  electricity use time series data. In: International conference on adaptive
  and natural computing algorithms, Springer, pp 401--412

\bibitem[{Ribeiro et~al.(2016)Ribeiro, Singh, and Guestrin}]{ribeiro2016should}
Ribeiro MT, Singh S, Guestrin C (2016) " why should i trust you?" explaining
  the predictions of any classifier. In: Proceedings of the 22nd ACM SIGKDD
  international conference on knowledge discovery and data mining, pp
  1135--1144

\bibitem[{Rousseeuw(1987)}]{rousseeuw1987silhouettes}
Rousseeuw PJ (1987) Silhouettes: a graphical aid to the interpretation and
  validation of cluster analysis. Journal of computational and applied
  mathematics 20:53--65

\bibitem[{Schlegel et~al.(2019)Schlegel, Arnout, El-Assady, Oelke, and
  Keim}]{schlegel2019towards}
Schlegel U, Arnout H, El-Assady M, Oelke D, Keim DA (2019) Towards a rigorous
  evaluation of xai methods on time series. In: 2019 IEEE/CVF International
  Conference on Computer Vision Workshop (ICCVW), IEEE, pp 4197--4201

\bibitem[{Selvaraju et~al.(2017)Selvaraju, Cogswell, Das, Vedantam, Parikh, and
  Batra}]{selvaraju2017grad}
Selvaraju RR, Cogswell M, Das A, Vedantam R, Parikh D, Batra D (2017) Grad-cam:
  Visual explanations from deep networks via gradient-based localization. In:
  Proceedings of the IEEE international conference on computer vision, pp
  618--626

\bibitem[{Shrikumar et~al.(2017)Shrikumar, Greenside, and
  Kundaje}]{deepliftshrikumar2017learning}
Shrikumar A, Greenside P, Kundaje A (2017) Learning important features through
  propagating activation differences. In: Proceedings of the 34th International
  Conference on Machine Learning-Volume 70, JMLR. org, pp 3145--3153

\bibitem[{Wang et~al.(2017)Wang, Yan, and Oates}]{wang2016time}
Wang Z, Yan W, Oates T (2017) Time series classification from scratch with deep
  neural networks: A strong baseline. In: 2017 International Joint Conference
  on Neural Networks (IJCNN), pp 1578--1585

\end{thebibliography}


\end{document}